\documentclass[letterpaper]{article} 
\usepackage{aaai2027}  
\usepackage[hyphens]{url}  
\usepackage{graphicx} 
\usepackage{natbib}  
\usepackage{caption} 
\usepackage{algorithm}
\usepackage{algorithmicx}
\usepackage{algpseudocode}

\usepackage{newfloat}
\usepackage{listings}
\DeclareCaptionStyle{ruled}{labelfont=normalfont,labelsep=colon,strut=off} 
\floatstyle{ruled}
\newfloat{listing}{tb}{lst}{}
\floatname{listing}{Listing}

\usepackage{booktabs}

\usepackage{amsmath}
\usepackage{amsfonts}

\def\our{PiPS}

\title{\our{}: Post-Hoc Prototypical Explanations for Interpretable Semantic Segmentation}
\author{
    Miłosz Adamczyk\textsuperscript{\rm 1},
    Tymoteusz Zapala \textsuperscript{\rm 2},
    Piotr Borycki\textsuperscript{\rm 1, 3},
    Przemys{\l}aw Spurek\textsuperscript{\rm 1, 3},
}
\affiliations{
    \textsuperscript{\rm 1} Jagiellonian University\\
    \textsuperscript{\rm 2} Wrocław University of Science and Technology\\
    \textsuperscript{\rm 3} IDEAS Research Institute \\
}

\begin{document}

\maketitle

\begin{abstract}
With the increasing deployment of deep neural networks in critical systems, such as medical diagnostics and autonomous vehicles, ensuring their interpretability is crucial to building trust in decision-making systems. In the field of explainable artificial intelligence, prototype-based reasoning has gained particular popularity, as it mimics human cognitive processes by explaining model decisions based on visual similarity under the \textit{looks like this} paradigm. While this paradigm has been thoroughly investigated in the context of global image classification, the interpretability of dense predictions, particularly semantic segmentation, remains largely unexplored despite its immense importance in tasks requiring precise object localization. Existing prototype-based interpretable segmentation models rely on ante-hoc architectures, which entails significant limitations because they require costly training from scratch and modifications to the network structure, ultimately leading to a noticeable drop in predictive performance compared to standard black-box models. To address this issue, we propose \our{} (Post-hoc interpretable Prototypical Segmentation), the first fully post-hoc solution for generating prototypical explanations for semantic segmentation models. Our method enables the extraction of intuitive, spatially localized explanations from any pre-trained network without modification or fine-tuning, thereby preserving 100\% of the model's original predictive performance. This approach opens a new avenue for the safe and cost-effective deployment of transparent systems in advanced computer vision tasks. Codebase available at \url{https://github.com/gmum/PIPS}
\end{abstract}


\begin{figure*}[t]
    \centering
    \includegraphics[width=\linewidth]{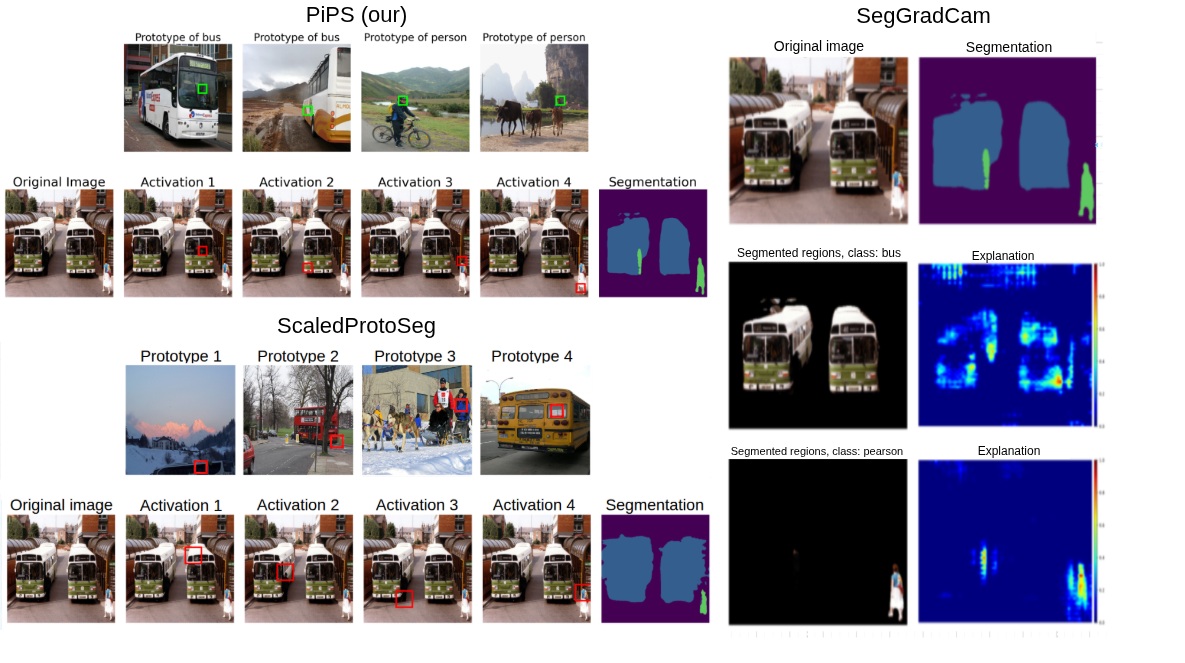}
    \caption{\textbf{Comparison of explanation methods on a multi-object scene (bus and person) from PASCAL VOC 2012.} \textbf{PiPS (Ours)} delivers a disentangled, part-by-part decomposition of the object, explicitly visualizing the local features that influenced the model's decision by comparing them against training data. Every two rows correspond to a specific prototypical part (showing the top-2 channels for each predicted class). Green bounding boxes highlight the activation of a given part within the reference training images, while red boxes in the leftmost column pinpoint the corresponding parts in the original target test image (the second column displays segmentation masks).\textbf{ScaledProtoSeg} yields similarly looking results and outputs a fixed number of prototypical parts arbitrarily chosen regardless of the actual number of classes present. Similarly, \textbf{SegGradCam} produces general-purpose saliency maps rather than explicit part activations, but provides exactly one attribution map per detected class.}
    \label{fig:method_comparison}
\end{figure*}

\begin{figure}[htbp]
    \centering
    \includegraphics[width=\linewidth, keepaspectratio]{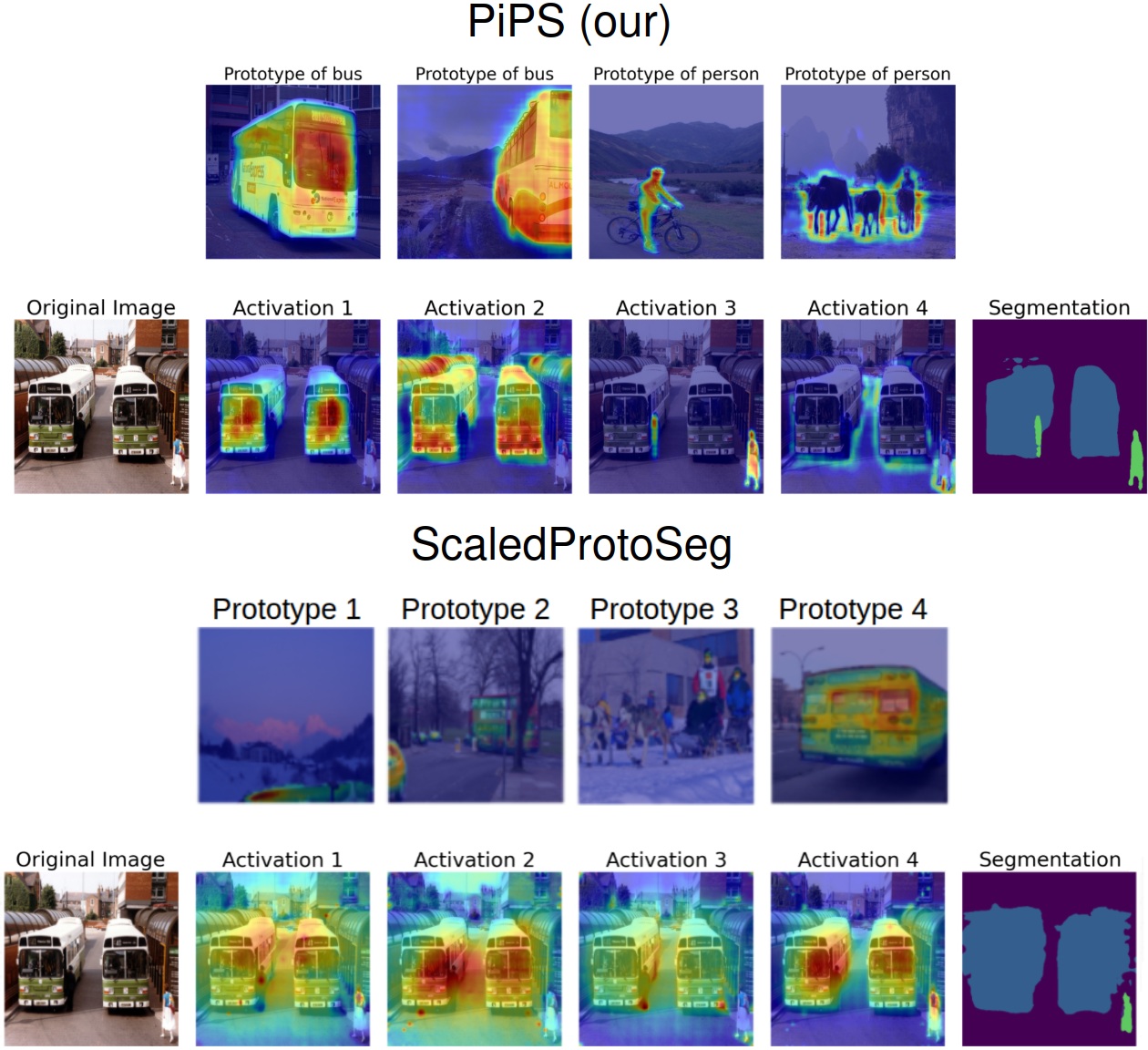}
    \caption{\textbf{Comparison of PiPS and ScaledProtoSeg with heatmaps.} Visual evaluation using heatmaps illustrating the spatial activation patterns and localization performance of prototypical parts generated by PiPS and ScaledProtoSeg on semantic segmentation tasks. Notice the variations in activation in compared approaches.}
    \label{fig:pips_vs_scaledprotoseg}
\end{figure}

\section{Introduction}
\label{sec:introduction}

As deep neural networks continue to achieve unprecedented performance across a multitude of computer vision tasks, their inherent black-box nature poses a significant barrier to their deployment in real-world scenarios. Ensuring model interpretability has thus become a fundamental requirement for building trust, accountability, and safety in artificial intelligence systems. Within the domain of Explainable Artificial Intelligence (XAI), prototype-based reasoning has emerged as a leading and highly intuitive concept. By adopting the \textit{looks like this} paradigm~\cite{chen2019looks}, prototypical methods mimic human cognitive processes, explaining a model's decision by pointing out visual similarities between parts of an input image and a set of learned, representative prototypes.

While prototypical explainability has been extensively studied and successfully implemented in the context of global image classification~\cite{chen2019looks}, its application to dense prediction tasks remains largely unexplored. This constitutes a critical gap in the literature, given that semantic segmentation provides the fine-grained, pixel-level scene understanding indispensable for numerous high-stakes applications. In domains such as autonomous driving, medical image analysis, and robotic navigation, merely knowing that a specific class is present in an image is insufficient. In these fields, precise spatial localization and local interpretability are paramount, making the ability to explain exactly why a specific group of pixels was assigned to a given class a crucial safety requirement.

Despite the pressing need for interpretable dense predictions, existing attempts to bring prototypical reasoning to semantic segmentation rely almost exclusively on ante-hoc architectures~\cite{sacha2023protoseg, porta2025multi}. These approaches embed custom prototype layers directly into the network design, which imposes severe limitations. Primarily, they require computationally expensive training from scratch, making it exceptionally difficult to adapt them to modern, massive foundation models. Furthermore, they introduce a problematic trade-off: forcing the network to learn a constrained, prototype-friendly latent space typically leads to a noticeable degradation in predictive performance, resulting in a lower mean Intersection over Union (mIoU) compared to state-of-the-art black-box baselines. On the other hand, while post-hoc methods have been adapted for segmentation, they are largely limited to attribution maps or adversarial frameworks~\cite{gipivskis2025post, selvaraju2017grad}, which inherently lack the human-understandable visual reasoning that prototypes provide.

To overcome these fundamental limitations, we propose \our{} (Post-hoc interpretable Prototypical Segmentation), the first fully post-hoc solution designed specifically for prototypical semantic segmentation. Instead of modifying and retraining the network, \our{} operates on top of existing, pre-trained models. Our method works by extracting dense, pixel-wise feature maps from the deep layers of a frozen base model and mathematically projecting them onto a set of representative spatial prototypes. By decoupling the explanation mechanism from the primary prediction pipeline, \our{} successfully translates the complex latent representations of dense tasks into intuitive, region-based visual similarities without interfering with the model's original weights (see Figure~\ref{fig:method_comparison} for a visual comparison of our explanations against existing baseline methods and Figure~\ref{fig:pips_vs_scaledprotoseg} for comparsion using heatmaps).

The main contributions of our work can be summarized as follows:
\begin{itemize}
    \item We introduce \our{}, the first fully post hoc framework for prototype-based explainability in semantic segmentation, which eliminates the need for architectural modifications and costly retraining from scratch.
    \item We achieve complete preservation of the base model's predictive performance, effectively breaking the accuracy-versus-interpretability trade-off that severely limits existing ante hoc prototypical segmentation methods.
    \item We demonstrate the broad flexibility and universal applicability of our approach by successfully integrating it with various state-of-the-art pre-trained segmentation networks, providing highly localized, human-understandable visual explanations across complex datasets.
\end{itemize}

\begin{figure*}[t]
    \centering
    \includegraphics[width=\textwidth]{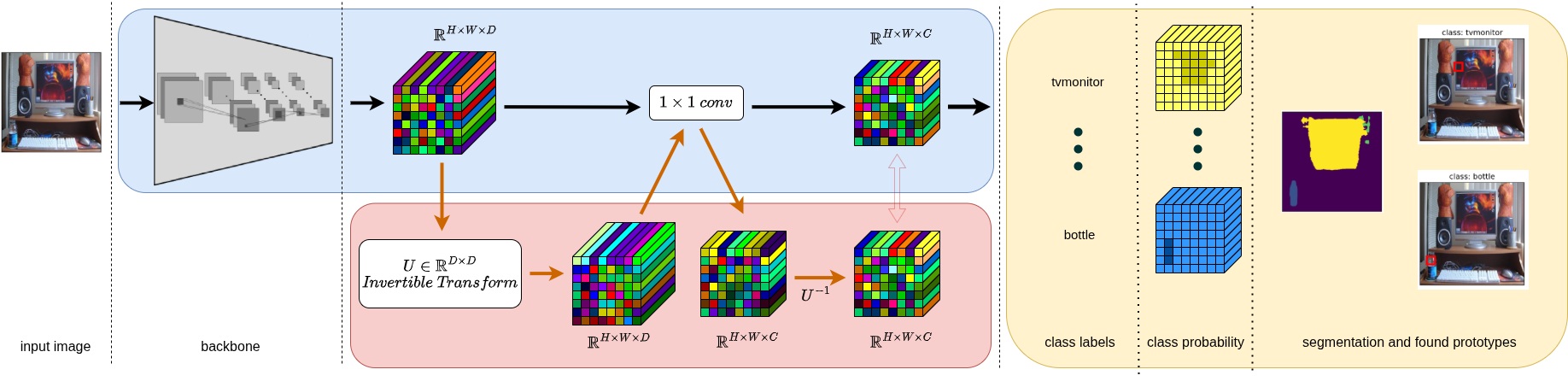}
    \caption{\textbf{Overview of the PiPS training framework.} The pre-trained backbone features are passed through a trainable invertible transformation matrix. We optimize a purity objective to isolate localized concepts. During training, we dynamically subset the dataset based on exemplar scores to progressively focus on the most representative prototypes.}
    \label{fig:training_process}
\end{figure*}


\section{Related Works}
\label{sec:related_works}

With the rapid development and increasingly widespread deployment of deep learning models in key areas such as healthcare and autonomous systems, the issue of explainability has become a fundamental research challenge. In the scholarly literature on explainable artificial intelligence (XAI), two principal paradigms can be distinguished: post hoc explanation methods and inherently interpretable (ante hoc) models.

Post hoc methods focus on analyzing already-trained models, providing explanations without altering their architecture. Classic examples include SHAP~\cite{lundberg2017unified} and LIME~\cite{ribeiro2016should}, which assign importance to individual features, as well as Grad-CAM~\cite{selvaraju2017grad}, which generates attention maps highlighting critical input regions. Recently, the post-hoc approach has also been extensively explored in the context of semantic image segmentation, offering various attribution maps and even providing frameworks for adversarial attacks~\cite{gipivskis2025post}. However, despite their flexibility, post-hoc attribution methods are often criticized for the instability of the generated explanations~\cite{adebayo2018sanity} and their inability to provide intuitive, human-like visual reasoning.

To address the limitations of post-hoc attribution maps, ante-hoc models integrate interpretability mechanisms directly into the architecture. A prominent development in this area is the ProtoPNet algorithm~\cite{chen2019looks}, which introduces class prototypes, enabling interpretation of model decisions through the \textit{looks like this} paradigm. While originally designed for global image classification, this prototypical reasoning has recently been adapted to dense prediction tasks, where spatial localization is paramount. For instance, ProtoSeg~\cite{sacha2023protoseg} introduced interpretable semantic segmentation using prototypical parts, enabling models to explain pixel-wise decisions based on local visual similarities. This concept was further advanced by introducing multi-scale grouped prototypes~\cite{porta2025multi}, which allow the network to capture and explain objects at various spatial granularities. 

Despite providing highly intuitive and localized explanations, prototypical segmentation models face a critical bottleneck, they are strictly ante-hoc. Methods such as ProtoSeg~\cite{sacha2023protoseg} and multi-scale prototype networks~\cite{porta2025multi} require specialized architectures and computationally expensive training from scratch. This makes them difficult to apply to modern, large-scale foundation models. Furthermore, forcing the network to learn a constrained, prototype-based latent space typically leads to a noticeable degradation in predictive performance (e.g., lower mean Intersection over Union) compared to state-of-the-art black-box models. 

While hybrid methods that combine post-hoc flexibility with concept-based explanations, such as ACE~\cite{ghorbani2019automating} or Concept Whitening~\cite{chen2020concept}, have been proposed for classification, they fail to provide the exact, visually verifiable prototypical patches that are highly desired in dense tasks. Recently, EPIC~\cite{EPIC2026} introduced a post-hoc prototype framework for global image classification. However, EPIC is inherently designed for whole-image classification and lacks the spatially bounded attribution mechanism required to explain dense, pixel-level prediction tasks. Thus, a clear gap exists: the literature lacks a method that provides the deep interpretability of prototypical parts combined with the flexibility of post-hoc application specifically tailored for semantic segmentation.

Our proposed method, \our{}, addresses this exact gap by enabling prototype-based explanations on top of already trained segmentation networks. It successfully combines the scalability offered by post hoc techniques~\cite{gipivskis2025post} with the localized interpretability characteristic of ante hoc prototypical segmentation approaches~\cite{sacha2023protoseg, porta2025multi}. Crucially, \our{} achieves this without requiring any architectural modifications or retraining, thereby preserving 100\% of the original model's performance.

\begin{figure*}[htbp] 
    \centering
    \includegraphics[width=\textwidth, keepaspectratio]{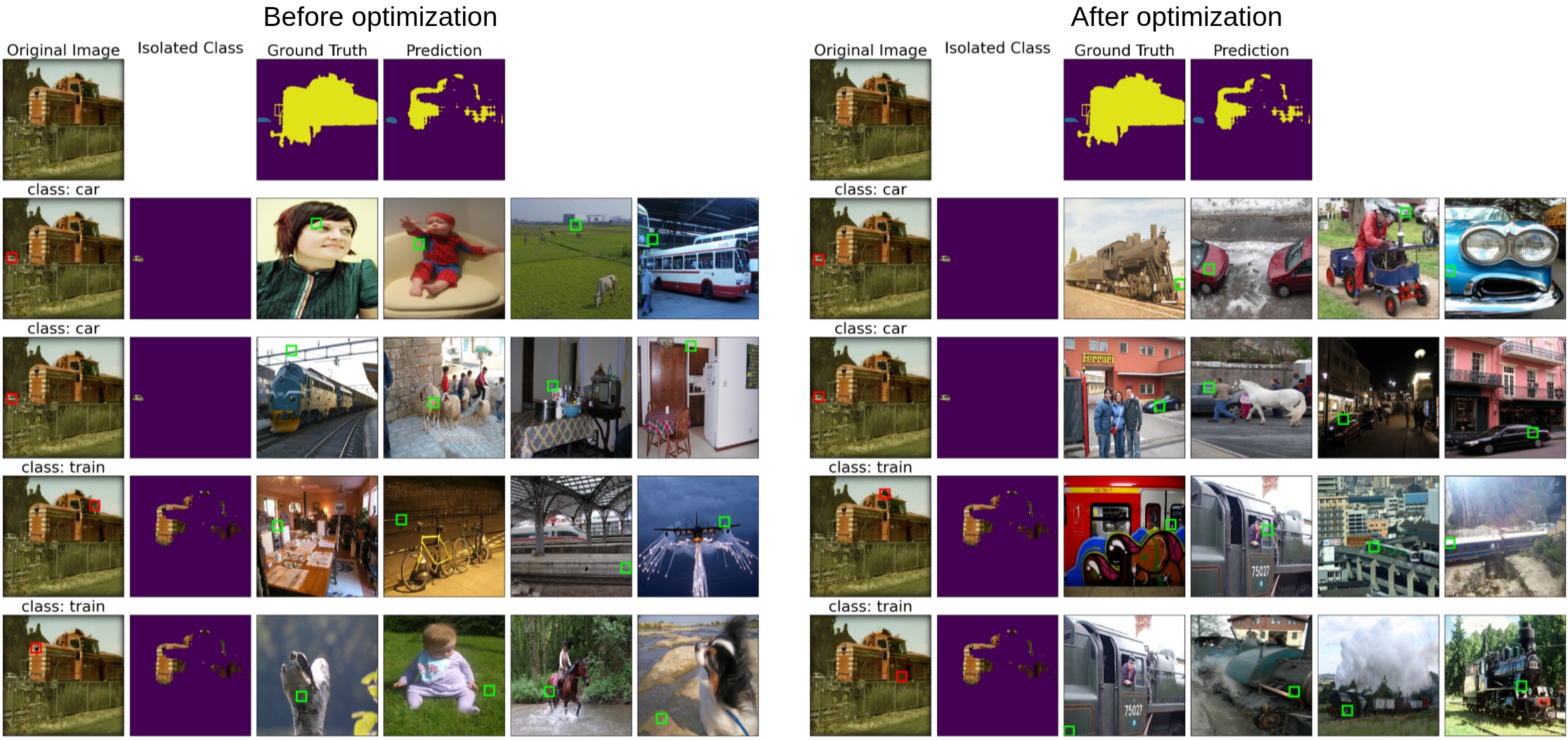}
    \caption{\textbf{Impact of the optimization phase on the generated explanations.} Visual comparison of the extracted prototypical parts before (left) and after (right) the purity-driven optimization process on the pre-trained segmentation backbone. As observed, prototypes without the additional tuning correspond to random, uninterpretable image patches with limited explanatory properties. After the optimization, these prototypes become highly consistent and accurately correspond with the localized semantic features of the input image.}
\label{fig:3beforeafterepic}
\end{figure*}

\section{\our{}}
In this section, we present our post-hoc interpretability framework for explaining pre-trained semantic segmentation networks. While global model-attribution methods are typically designed to identify whole-image-level concepts for classification tasks, explaining dense pixel-level predictions requires strict spatial localization. To address this challenge, our framework leverages a post-hoc orthogonal coordinate rotation, optimizes a spatially-formulated purity objective to isolate localized concepts, and utilizes a segment-constrained attribution mechanism to ensure that explanations are structurally bound to the corresponding predicted class boundaries (an overview of our training pipeline is illustrated in Figure \ref{fig:training_process}).

\paragraph{Orthogonal Transformation} To interpret the network's latent representations without modifying its underlying feature-extraction capabilities, we freeze all backbone parameters of the pre-trained segmentation network and modify only the final $1 \times 1$ convolutional classifier layer of the segmentation head. Let $W \in \mathbb{R}^{C_{\text{out}} \times C_{\text{in}} \times 1 \times 1}$ and $b \in \mathbb{R}^{C_{\text{out}}}$ denote the pre-trained weights and bias of the classifier, where $C_{\text{in}}$ and $C_{\text{out}}$ represent the input channel dimension and the number of target semantic classes, respectively. We introduce a trainable, raw parameter matrix $A \in \mathbb{R}^{C_{\text{in}} \times C_{\text{in}}}$. An orthogonal transformation matrix $M \in \mathbb{R}^{C_{\text{in}} \times C_{\text{in}}}$ is parameterized via the matrix exponential of the skew-symmetric matrix derived from $A$:
\begin{equation*}
A = A - A^T, \quad M = \exp(A)
\end{equation*}
The skew-symmetry of $A$ guarantees the mathematical orthogonality of $M$. For a latent activation tensor $X \in \mathbb{R}^{B \times C_{\text{in}} \times H \times W}$ prior to the classifier, the disentangled representations $Z \in \mathbb{R}^{B \times C_{\text{in}} \times H \times W}$ are computed by applying the coordinate rotation along the channel dimension:
\begin{equation*}
Z = X M
\end{equation*}
To maintain exact mathematical equivalence with the original classifier mapping, the pre-trained weights are rotated dually:
\begin{equation*}
W_{\text{new}} = W M
\end{equation*}
The final classification logits are subsequently computed as $\hat{Y} = \text{Conv2d}(Z, W_{\text{new}}, b)$, ensuring that the transformation remains strictly post-hoc and preserves the original predictions of the network.

\paragraph{Optimization Objective} 
To align the rotated latent coordinates in $Z$ with discrete, human-interpretable semantic concepts, we train the parameter $A$ using a spatially-formulated purity loss. For each channel $c \in \{1, \dots, C_{\text{in}}\}$ of a sample $b$ in a batch, the spatial activation map $z_{b, c} \in \mathbb{R}^{N}$ (where $N = H \times W$) is converted to a spatial probability distribution using the softmax operator:
\begin{equation*}
p_{b, c, i} = \frac{\exp(z_{b, c, i})}{\sum_{j=1}^{N} \exp(z_{b, c, j})}
\end{equation*}
We define the spatial entropy $H(p_{b, c})$ to measure the dispersion of the channel's activation:
\begin{equation*}
H(p_{b, c}) = -\sum_{i=1}^{N} p_{b, c, i} \log(p_{b, c, i} + \epsilon)
\end{equation*}
where $\epsilon = 10^{-8}$. To focus the optimization on highly active channels while discounting dead or noisy features, we weight each channel's entropy by its normalized peak absolute activation:
\begin{equation*}
a_{b, c} = \max_{i} |z_{b, c, i}|, \quad w_{b, c} = \frac{a_{b, c}}{\sum_{k=1}^{C_{\text{in}}} a_{b, k} + \epsilon}
\end{equation*}
The objective function is then defined as the expectation of the weighted spatial entropies:
\begin{equation*}
\mathcal{L}_{\text{purity}} = \frac{1}{B} \sum_{b=1}^{B} \sum_{c=1}^{C_{\text{in}}} w_{b, c} H(p_{b, c})
\end{equation*}
Minimizing $\mathcal{L}_{\text{purity}}$ guides $M$ to align the coordinate space such that individual latent channels capture highly localized, compact spatial regions. The qualitative results obtained before and after this optimization phase can be observed in Figure \ref{fig:3beforeafterepic}.

\paragraph{Training Strategy}
We train the disentanglement parameters for a total of $E = 20$ epochs. To systematically filter out noisy spatial activations and accelerate convergence, every two epochs, the training dataset is dynamically subset. For every image, we calculate a purity-based exemplar score $S(b, c)$ for each channel:
\begin{equation*}
S(b, c) = \frac{\max_{i} |z_{b, c, i}|}{H(p_{b, c}) + \epsilon}
\end{equation*}
For each channel $c$, we identify the top $K$ training samples that yield the highest exemplar scores $S(b, c)$. To progressively narrow the training distribution to the most representative prototypes, we linearly decay $K$ from 100 to 5 over the course of training.

\paragraph{Segment Visualization}
During inference, we generate visual explanations by identifying the training prototypes that correspond to active latent channels. While original image-level attribution models evaluate features globally, our semantic segmentation setting demands that explanations be spatially bounded to the exact predicted class boundaries to avoid background contamination. Given a test image, let $\hat{Y}_{\text{mask}}$ be the model's categorical segmentation output. For a target class $c_{\text{pred}}$ within the prediction, we define a binary spatial mask $M_{c_{\text{pred}}} = \mathbb{I}(\hat{Y}_{\text{mask}} = c_{\text{pred}})$. We resize $M_{c_{\text{pred}}}$ to match the spatial dimensions of the latent features using nearest-neighbor interpolation, yielding $M^{\text{resized}}_{c_{\text{pred}}}$. For each active channel $c$, we restrict the feature evaluation strictly to this predicted segment:
\begin{equation*}
z'_{c, i} = \begin{cases} 
|z_{c, i}| & \text{if } i \in M^{\text{resized}}_{c_{\text{pred}}} \\ 
-1.0 & \text{otherwise} 
\end{cases}
\end{equation*}
We then identify the peak spatial activation coordinate within the segment:
\begin{equation*}
i^* = \arg\max_{i} z'_{c, i}
\end{equation*}
A visual bounding patch of size $P \times P$ is centered around the projected coordinate of $i^*$ on the input image. This step ensures that the attribution is strictly localized within the segment boundary of the predicted class. The resulting test patches are compared with the top-scoring training exemplars to provide localized, human-interpretable visual evidence for the model's pixel-level decisions.

\paragraph{Adaptation to Point Clouds}
Although our framework is presented in the context of image semantic segmentation, it is not restricted to the image domain. The only requirement imposed by \our{} is that the underlying segmentation architecture produces localized latent representations that correspond to spatially coherent regions and are subsequently used for dense prediction. This property is naturally satisfied by transformer-based point cloud segmentation models, such as the Point Transformer architecture~\cite{zhao2021pointtransformer}, where the input point cloud is first partitioned into local groups. Given an input point cloud $P \in \mathbb{R}^{N \times 3}$, Farthest Point Sampling (FPS) selects a set of $G$ representative group centers,
\[
C=\mathrm{FPS}(P,G),
\]
after which each center is assigned its $K$ nearest neighboring points using the K-Nearest Neighbors (KNN) algorithm,
\[
\mathcal{G}_g=\mathrm{KNN}(c_g,K), \qquad g=1,\ldots,G.
\]
The resulting local groups are processed by the Point Transformer backbone, yielding patch-level latent representations
\[
Z=f_{\mathrm{PT}}(\{\mathcal{G}_g\}_{g=1}^{G}).
\]
Following Point-BERT~\cite{yu2022pointbertpretraining3dpoint} and Point-MAE~\cite{pang2022maskedautoencoderspointcloud}, the representations from the final transformer layer are propagated back to the original point cloud using the PointNet++~\cite{qi2017pointnetdeephierarchicalfeature} feature propagation module, where each point feature is reconstructed by interpolating the embeddings of its nearest patch centers before point-wise classification. Analogously to the image setting, we insert the invertible orthogonal transformation immediately after the final patch representations and compensate it by rotating the subsequent classifier weights, thereby preserving the original logits exactly. Consequently, the same purity objective, training strategy, and prototype retrieval procedure described above can be applied without modification, with the only distinction being that prototypes correspond to localized point groups instead of image patches.

\section{Experiments and Results}

\begin{table}[t]
\centering
\caption{Segmentation accuracy (mIoU \%) on the PASCAL VOC 2012 validation dataset. We compare our post-hoc method (PiPS) and SegGradCam against ante-hoc prototypical models (ProtoSeg~\cite{sacha2023protoseg}, ScaleProtoSeg~\cite{porta2025multiscale}). Notice that while ante-hoc modifications noticeably degrade the predictive performance of their DeepLabV2 baseline, post-hoc methods preserve 100\% of the original DeepLabV3 black-box accuracy, successfully breaking the accuracy-versus-interpretability trade-off.}
\label{tab:performance_comparison}
\vspace{0.1cm}
\resizebox{\columnwidth}{!}{%
\begin{tabular}{llc}
\toprule
\textbf{Method} & \textbf{Backbone} & \textbf{mIoU} \\
\midrule
Original Baseline & DeepLabV2 & 77.69\%~\cite{porta2025multiscale} \\
ProtoSeg & DeepLabV2 & 71.98\%~\cite{porta2025multiscale} \\
ScaleProtoSeg & DeepLabV2 & 71.80\%~\cite{porta2025multiscale} \\
\midrule
Original Baseline & DeepLabV3 & 85.70\%~\cite{chen2017rethinking} \\
SegGradCam & DeepLabV3 & 85.70\% \\
\textbf{PiPS (ours)} & \textbf{DeepLabV3} & \textbf{85.70\%} \\
\bottomrule
\end{tabular}%
}
\end{table}

\begin{figure}[t]
    \centering
    \includegraphics[width=0.95\columnwidth]{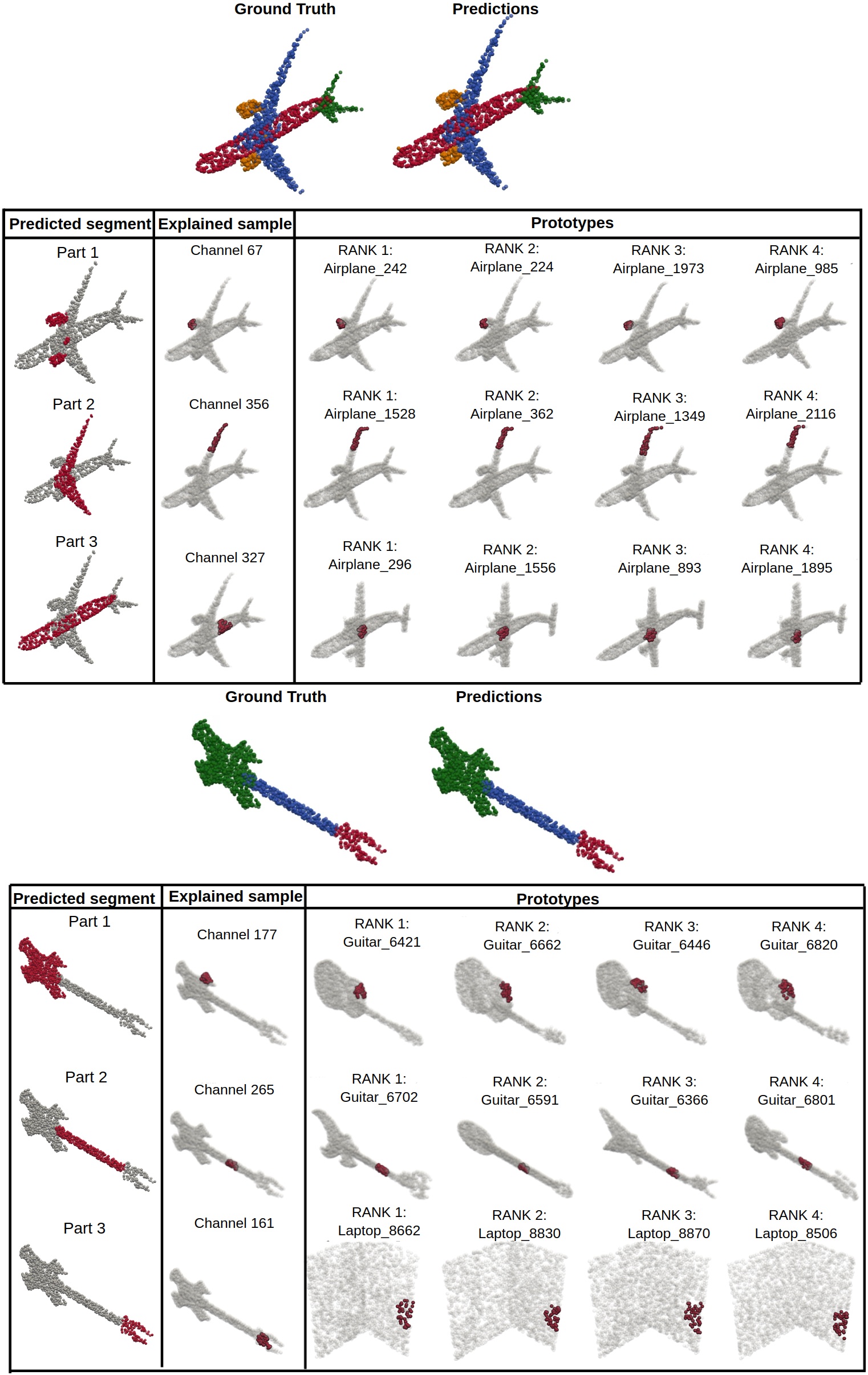}
    \caption{\textbf{Examples of a PiPS explanation for point cloud data.} For each predicted part class (first column) of the explained object, we select the patch with the most active channel (second column) that belongs to the corresponding segment. We then select prototypes from other objects for which the same channel is the most active (third column).}
    \label{fig:point-cloud-example}
\end{figure}

In the experimental section, we evaluate our \our{} framework across several scenarios. First, we provide a qualitative comparison, showcasing example predictions and comparing our results against the post-hoc method SegGradCam and the ante-hoc prototypical model ScaledProtoSeg. Then, we demonstrate that our method acts strictly as a post-hoc plugin, perfectly preserving the network's predictive performance without altering its original output. Finally, we present the structure and results of extensive user studies assessing the human-understandability of our approach.

\paragraph{Explanation of model decision} 
This section outlines the experimental results of \our{} explanations and its comparison to other XAI methods. Figure~\ref{fig:method_comparison}, illustrates the interpretability differences between \our{} and the classical post-hoc SegGradCam and the ante-hoc ScaledProtoSeg on multi-object scenes from the PASCAL VOC 2012 dataset. While SegGradCam produces diffused attribution maps that highlight general areas of importance, it falls short of capturing visually meaningful concepts such as textures or distinctive object parts. Similarly, ScaledProtoSeg struggles to provide clear, disentangled explanations without degrading the underlying predictions. In contrast, \our{} delivers a disentangled, part-by-part decomposition of the object. It not only highlights critical spatial regions but also explicitly pairs them with semantically rich training prototypes that represent these crucial visual features. Additional visual examples and comparisons can be found in the Appendix A6.

\paragraph{Segmentation Performance} 
As previously mentioned, the construction of \our{} preserves the predictive ability of the pre-trained backbone. This means that integrating the invertible transformation matrix does not change the model's categorical output. However, since we apply additional operations, minor numerical errors might theoretically arise. To demonstrate that this situation does not occur, we present the mean Intersection over Union (mIoU) numerical accuracy on the PASCAL VOC 2012 validation dataset in Table~\ref{tab:performance_comparison}. The results clearly show that while ante-hoc modifications (ProtoSeg, ScaledProtoSeg) noticeably degrade the predictive performance of their DeepLabV2 baseline, our post-hoc method preserves 100\% of the original DeepLabV3 black-box accuracy, successfully breaking the accuracy-versus-interpretability trade-off.

\paragraph{Point Cloud Segmentation}
To demonstrate that \our{} is not limited to image-based semantic segmentation, we additionally evaluate its point cloud adaptation on the ShapeNet Parts dataset~\cite{qi2017pointnetdeeplearningpoint}, which contains objects from 16 semantic categories, each composed of multiple part classes. The training procedure follows exactly the same optimization strategy as described for the image domain, with hyperparameters adjusted only to the structure of point cloud representations (see Appendix for implementation details). During inference, for every predicted part segment, we identify the latent channel exhibiting the strongest activation within the corresponding point group and retrieve the training object whose representation maximally activates the same channel. Using the hyperparameter configuration proposed by the original Point-BERT authors, the underlying segmentation model achieves an mIoU$_C$ of 84.1 and an mIoU$_I$ of 85.6, matching the performance reported for the original Point-BERT model. This confirms that the proposed post-hoc disentanglement preserves the model logits exactly and therefore does not alter the segmentation performance. Example of the generated point cloud prototypes are presented in Figure~\ref{fig:point-cloud-example}.

\begin{table*}[t]
\centering
\small
\caption{User study evaluation metrics ($N=26$). Overall ratings on a 5-point Likert scale (1-5) are reported alongside one-sample Wilcoxon signed-rank tests against the neutral midpoint (3.0) and Friedman tests for item variability.}
\label{tab:user_study_stats}
\begin{tabular}{lcccc}
\toprule
\textbf{Metric} & \textbf{Mean $\pm$ SD} & \textbf{Median} & \textbf{Wilcoxon $V$ ($p$ vs. 3.0)} & \textbf{Friedman $\chi^2$ ($p$)} \\
\midrule
Visual Similarity (Q1) & $3.15 \pm 1.08$ & $3.0$ & $3410.5$ ($p = 0.0429$) & $34.36$ ($p = 2 \cdot 10^{-6}$) \\
Visual Coherence (Q2) & $3.56 \pm 1.06$ & $4.0$ & $4917.5$ ($p = 3 \cdot 10^{-9}$) & $27.46$ ($p = 5 \cdot 10^{-5}$) \\
Feature Presence (Q3) & $3.60 \pm 1.01$ & $4.0$ & $5923.0$ ($p = 1 \cdot 10^{-10}$) & $17.96$ ($p = 0.0030$) \\
\bottomrule
\end{tabular}
\end{table*}

\begin{figure}[t]
    \centering
    \includegraphics[width=\linewidth]{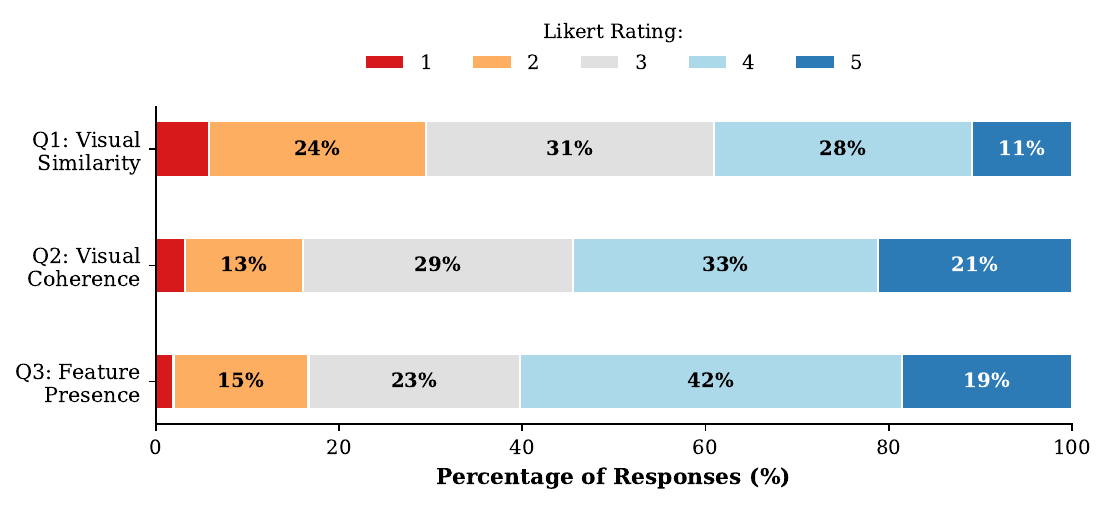}
    \caption{Percentage distribution of Likert scale ratings (1-5) for Visual Similarity, Visual Coherence, and Feature Presence ($N=26$).}
    \label{fig:likert_dist}
\end{figure}

\begin{figure}[t]
    \centering
    \includegraphics[width=\linewidth]{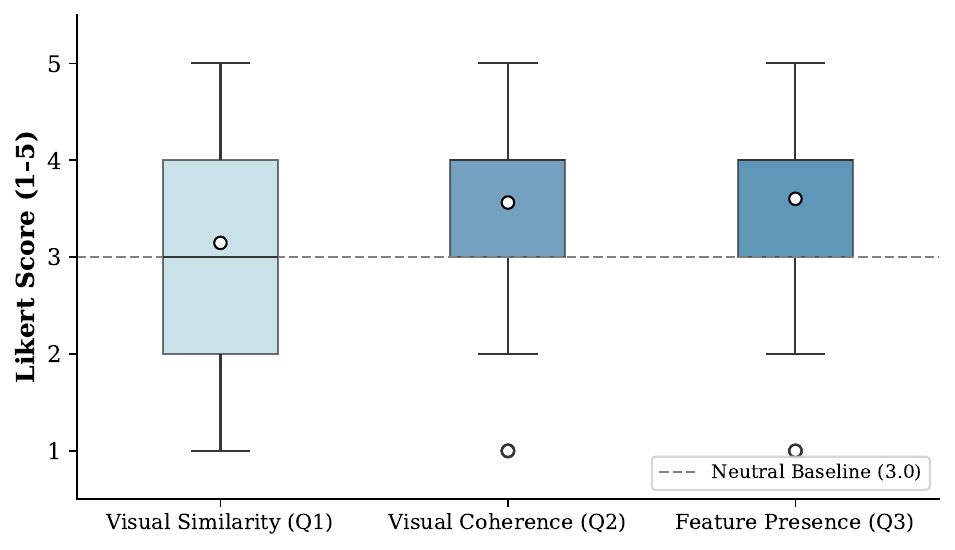}
    \caption{Boxplots showing Likert score distributions across question types. The white dot represents the mean score, and the dashed line denotes the neutral midpoint (3.0).}
    \label{fig:metrics_boxplot}
\end{figure}

\paragraph{User Study Results}
To comprehensively evaluate the human-understandability of our prototypical explanations, we conducted a user study ($N = 26$ participants, aged 20-50 with no specific background) utilizing dense multi-object scenes from the PASCAL VOC 2012 dataset. Participants evaluated the generated explanations on a 5-point Likert scale (1-5) across three core criteria: 
(i)~\textit{Visual Similarity} (extent to which the prototype matches visual elements in the input image), 
(ii)~\textit{Visual Coherence} (consistency of visual concepts across prototypes in a given row), and 
(iii)~\textit{Feature Presence} (verifiability of the prototype feature within the target image).

The quantitative results are summarized in Table~\ref{tab:user_study_stats}, with response distributions illustrated in Figures~\ref{fig:likert_dist} and~\ref{fig:metrics_boxplot}. Overall, participants rated \textit{Feature Presence} the highest ($\text{Mean} = 3.60 \pm 1.01$, $\text{Median} = 4.0$), closely followed by \textit{Visual Coherence} ($\text{Mean} = 3.56 \pm 1.06$, $\text{Median} = 4.0$). One-sample Wilcoxon signed-rank tests confirmed that both \textit{Feature Presence} ($V = 5923.0, p = 1 \cdot 10^{-10}$) and \textit{Visual Coherence} ($V = 4917.5, p = 3 \cdot 10^{-9}$) were rated statistically significantly higher than the neutral midpoint ($3.0$). \textit{Visual Similarity} achieved a positive score ($\text{Mean} = 3.15 \pm 1.08$, $\text{Median} = 3.0$), also significantly exceeding the neutral baseline ($V = 3410.5, p = 0.0429$). Statistically significant Friedman tests across all three metrics ($p \le 0.003$) indicate that user evaluations vary depending on the visual feature complexity of the underlying target images.

\paragraph{Limitations} Since the post-hoc coordinate rotation requires exact mathematical equivalence between the transformed features and the rotated classifier weights, it is primarily applicable to architectures with a linear or $1 \times 1$ convolutional classification head, making direct adaptation to non-linear or query-based decoders less straightforward.

\section{Conclusions}

In this work, we introduced \our{}, a fully post-hoc framework for prototype-based interpretability in semantic segmentation. By optimizing a spatially-formulated purity objective using an orthogonal coordinate transformation on frozen pre-trained backbones, \our{} extracts intuitive, region-localized, and human-understandable visual explanations without modifying network architectures or fine-tuning backbone weights. As a result, our method achieves 100\% preservation of the base model's original predictive performance, effectively resolving the performance-interpretability trade-off inherent to ante-hoc prototypical segmentation methods. Furthermore, we demonstrated the broad versatility of \our{} by successfully adapting it to 3D point cloud segmentation. Comprehensive qualitative evaluations, quantitative experiments, and user studies confirm the fidelity, clarity, and human-understandability of our prototypical explanations.

\bibliography{aaai2027}

\appendix
\setcounter{secnumdepth}{2}
\section{Appendix / Supplemental Material}

\subsection{Explanations of model decision}
In this section, we provide additional results of experiments regarding the explanations of model decisions made by PiPS, comparing it to both the post-hoc approach SegGradCam and the ante-hoc method ScaledProtoSeg. The experimental results, which highlight our method's ability to provide disentangled, part-by-part decomposition of objects across dense predictions, are presented on multi-object scenes from the PASCAL VOC 2012 in Fig. 3.

\subsection{More details on user study}
\label{subsec:user_study}
To evaluate the human-understandability of our prototypical explanations, we conducted a user study via the Google Forms platform. To ensure a diverse assessment, the survey questions were randomized, meaning each participant received a subset of unique queries. Data quality was strictly maintained by filtering out unengaged submissions---specifically, responses where participants selected the identical answer across all questions. In total, we collected 26 valid responses. Prior to the evaluation, participants were provided with comprehensive instructions and visual examples to familiarize themselves with the interpretation of our method's results.

Figures~\ref{fig:5.1userstudy},~\ref{fig:5.2userstudy} illustrate example questions used in the user study, where users were asked to evaluate:
\begin{itemize}
    \item The quality of the generated explanations.
    \item Which method generates better segmentations (only PiPS and ScaledProtoSeg were compared, since SegGradCam generates the same results as PiPS because of their post-hoc nature).
\end{itemize}

\subsection{Segmentation Performance}
As previously discussed, the strictly post-hoc design of PiPS maintains 100\% of the predictive performance of the pre-trained base model. In other words, integrating the invertible transformation matrix yields the exact same categorical segmentation output for any image as the original model. While additional operations could potentially introduce minor numerical errors, we demonstrate that this is not the case. The mIoU metric values are presented in table~\ref{tab:performance_comparison}. 

\begin{table}[htbp]
\centering
\caption{Segmentation accuracy (mIoU \%) on the PASCAL VOC 2012 validation dataset. We compare our post-hoc method (PiPS) and SegGradCam against ante-hoc prototypical models (ProtoSeg~\cite{sacha2023protoseg}, ScaleProtoSeg~\cite{porta2025multiscale}). Notice that while ante-hoc modifications noticeably degrade the predictive performance of their DeepLabV2 baseline, post-hoc methods preserve 100\% of the original DeepLabV3 black-box accuracy, successfully breaking the accuracy-versus-interpretability trade-off.}
\label{tab:performance_comparison}
\vspace{0.1cm}
\resizebox{\columnwidth}{!}{%
\begin{tabular}{llc}
\toprule
\textbf{Method} & \textbf{Backbone} & \textbf{mIoU} \\
\midrule
Original Baseline & DeepLabV2 & 77.69\%~\cite{porta2025multiscale} \\
ProtoSeg & DeepLabV2 & 71.98\%~\cite{porta2025multiscale} \\
ScaleProtoSeg & DeepLabV2 & 71.80\%~\cite{porta2025multiscale} \\
\midrule
Original Baseline & DeepLabV3 & 85.70\%~\cite{chen2017rethinking} \\
SegGradCam & DeepLabV3 & 85.70\% \\
\textbf{PiPS (ours)} & \textbf{DeepLabV3} & \textbf{85.70\%} \\
\bottomrule
\end{tabular}%
}
\end{table}

\subsection{Dataset}
In our experiments, we utilized the PASCAL VOC 2012~\cite{pascal-voc-2012} dataset, which is frequently employed in dense prediction and segmentation model evaluations. Its dense multi-object scenes, severe occlusions, and high intra-class spatial variations pose significant challenges for prototype-based models. It is worth noting that only a few of the previous prototypical parts-based methods, namely ProtoSeg, and than ScaledProtoSeg, have been successfully evaluated on such complex scenes.

\subsection{PACAL VOC 2012 Experiments}
All experiments were conducted using the Athena GPU cluster. It is important to note that since PiPS operates entirely post-hoc, the underlying segmentation network remains strictly frozen and is not retrained. The computational overhead is therefore dedicated exclusively to a lightweight optimization phase: training the post-hoc orthogonal transformation matrix via the spatial purity objective, and algorithmically mining a set of representative spatial prototypes from the training data. The duration of this process is highly dependent on the size of the training set and the latent resolution of the base model. For the PASCAL VOC 2012 dataset, optimizing the transformation parameters and dynamically subsetting the exemplars takes approximately one hour on a single GPU node for the DeepLabV3 model.

\subsection{PASCAL VOC 2012 examplec}

\begin{figure*}[t!] 
    \centering
    \includegraphics[width=0.6\textwidth]{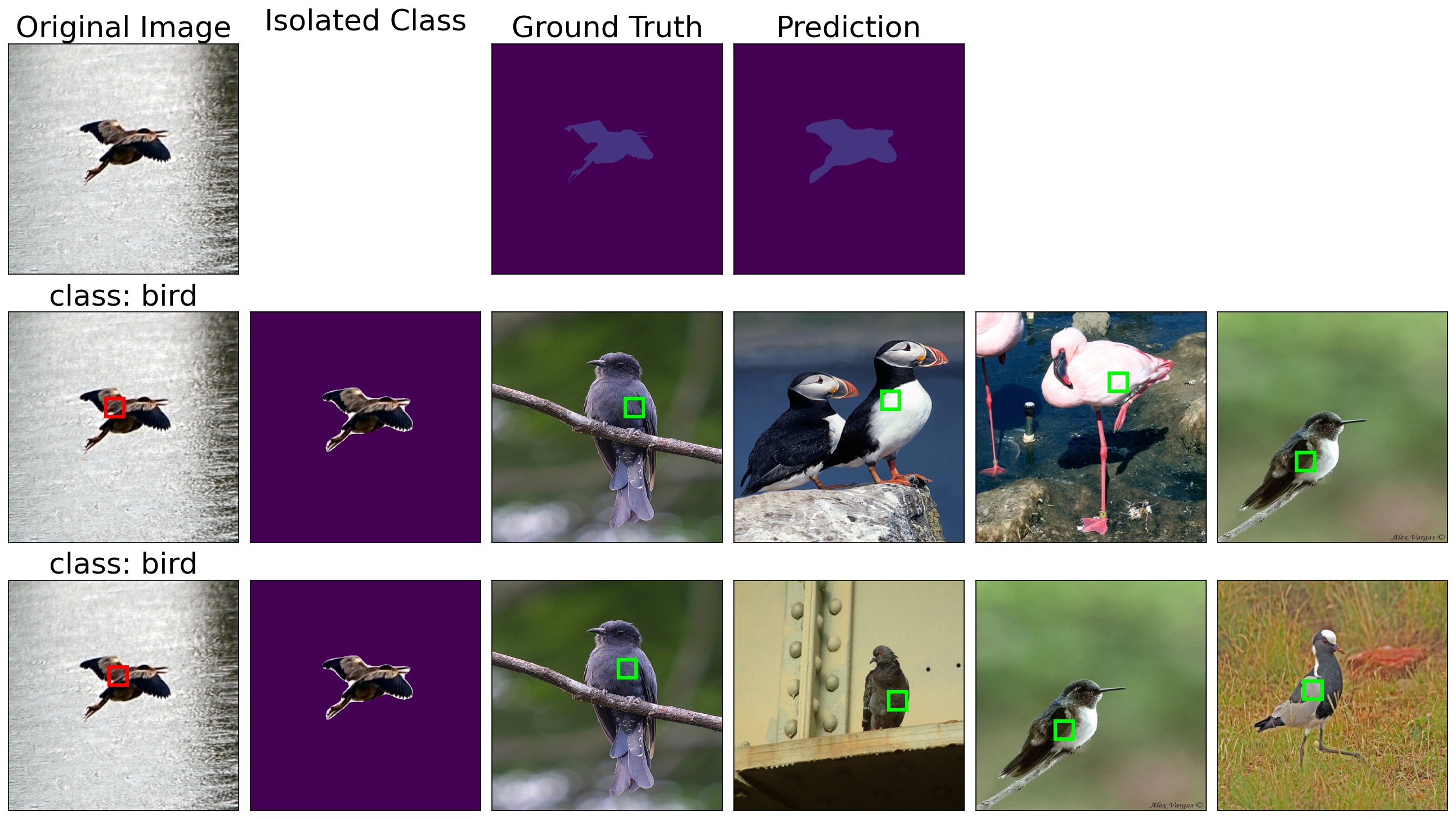}
    \caption{Visual explanation for the 'bird' object identified during semantic segmentation}
    \label{fig:example-PASCAL1}
\end{figure*}

\begin{figure*}[t!] 
    \centering
    \includegraphics[width=0.6\textwidth]{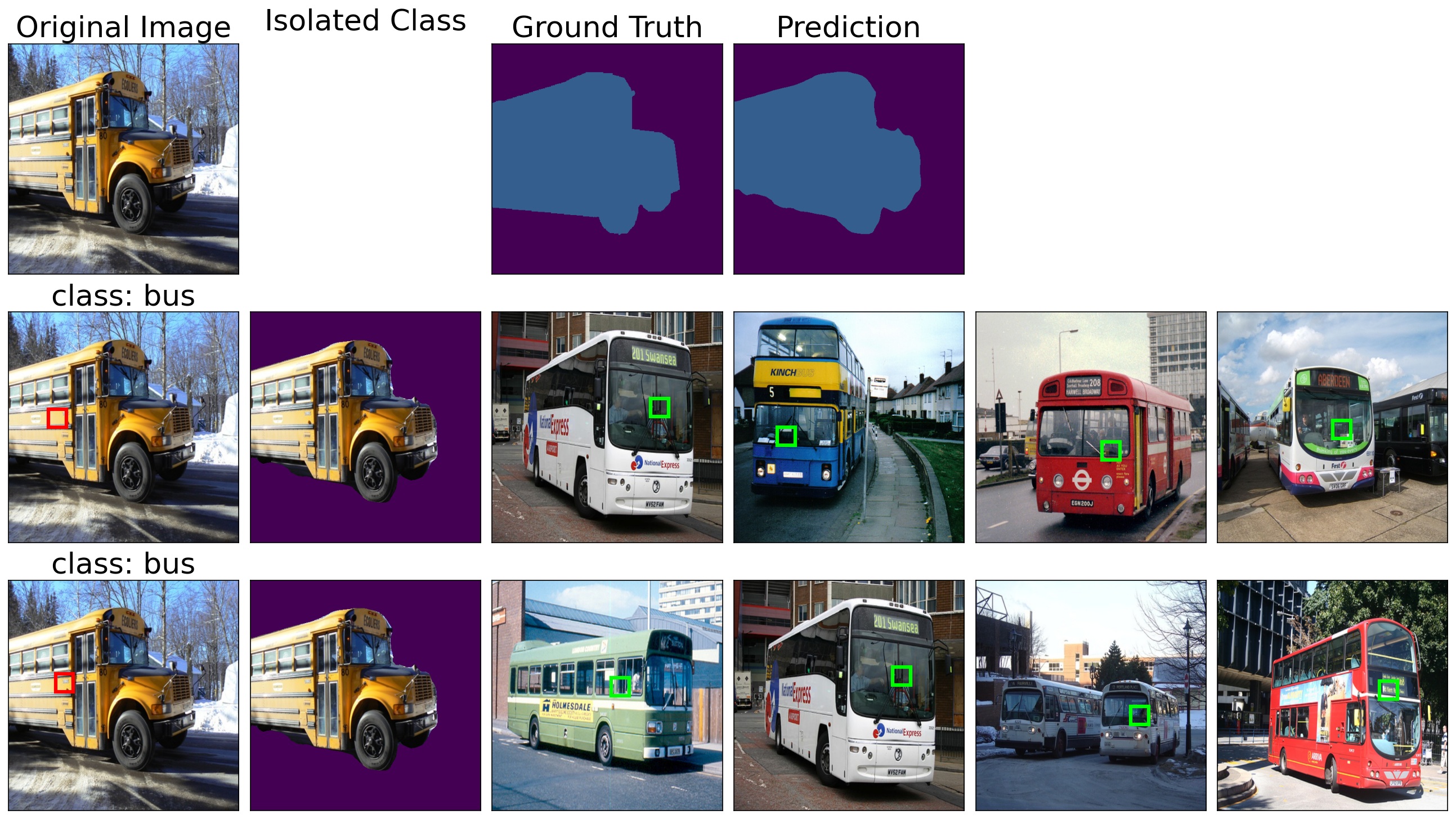}
    \caption{Visual explanation for the 'bus' object identified during semantic segmentation}
    \label{fig:example-PASCAL2}
\end{figure*}

\begin{figure*}[t!] 
    \centering
    \includegraphics[width=0.6\textwidth]{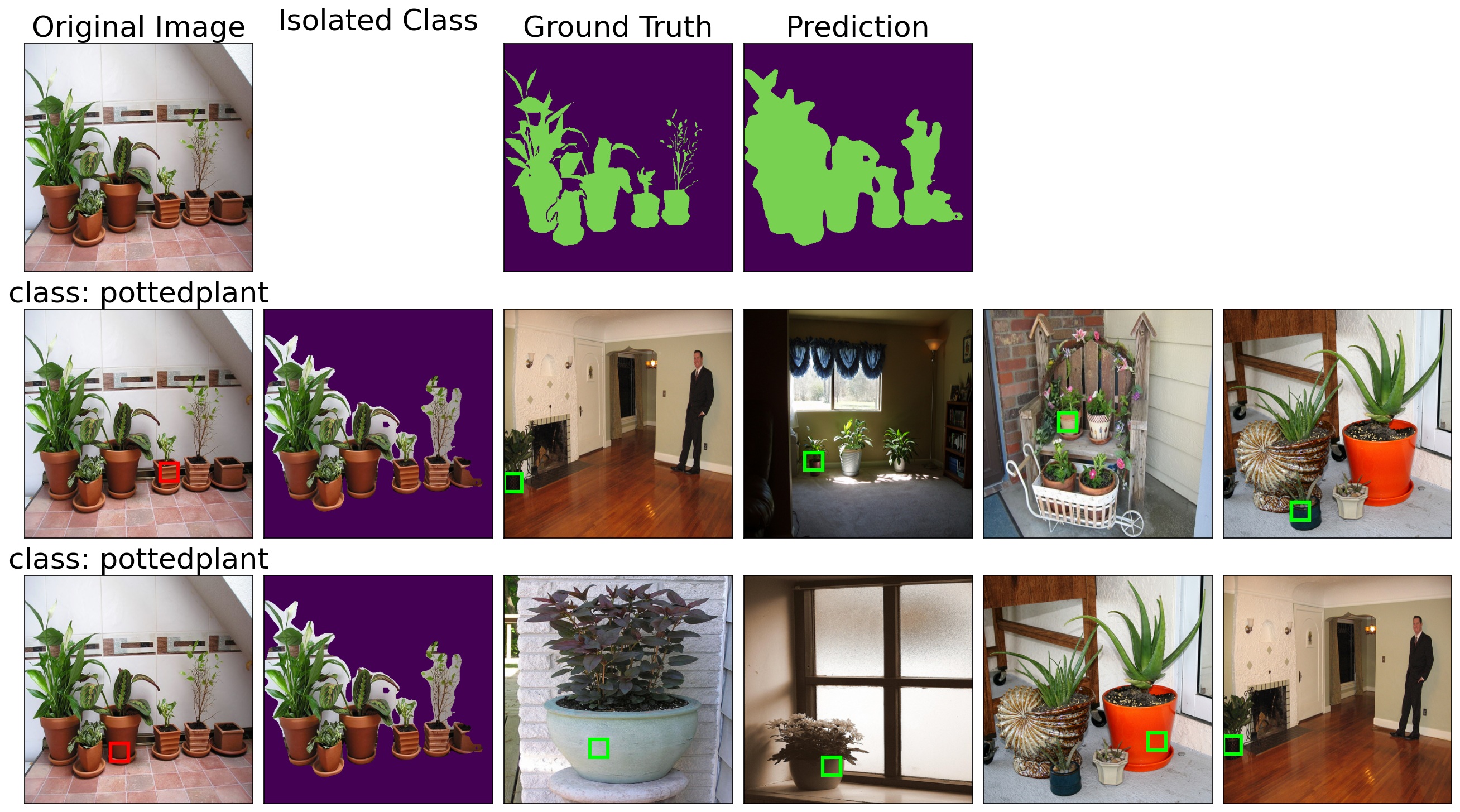}
    \caption{Visual explanation for the 'pottedplant' object identified during semantic segmentation}
    \label{fig:example-PASCAL3}
\end{figure*}

\begin{figure*}[t!] 
    \centering
    \includegraphics[width=0.6\textwidth]{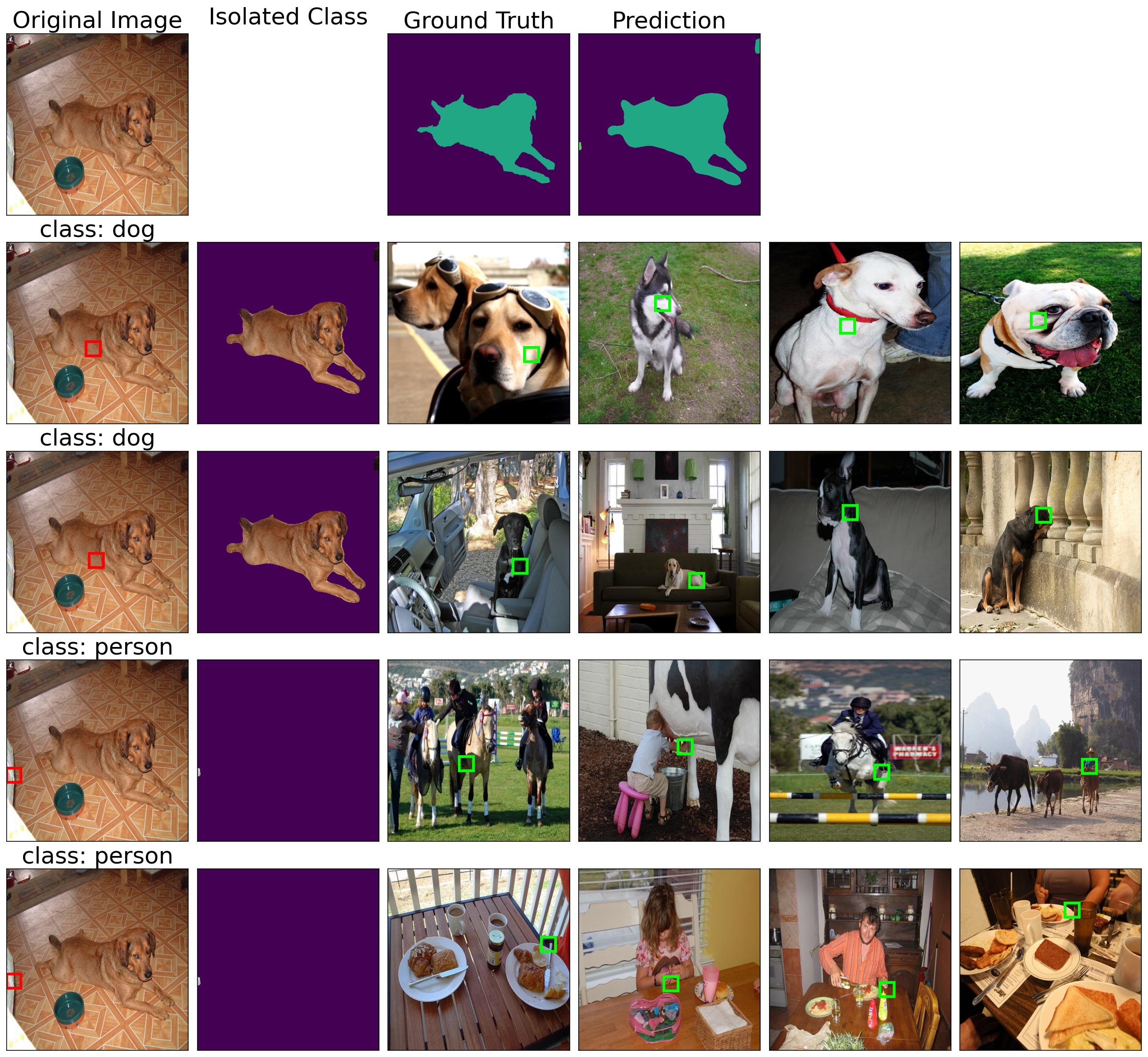}
    \caption{Visual explanation for the 'dog' and 'pearson' objects identified during semantic segmentation}
    \label{fig:example-PASCAL4}
\end{figure*}

\begin{figure*}[t!] 
    \centering
    \includegraphics[width=0.6\textwidth]{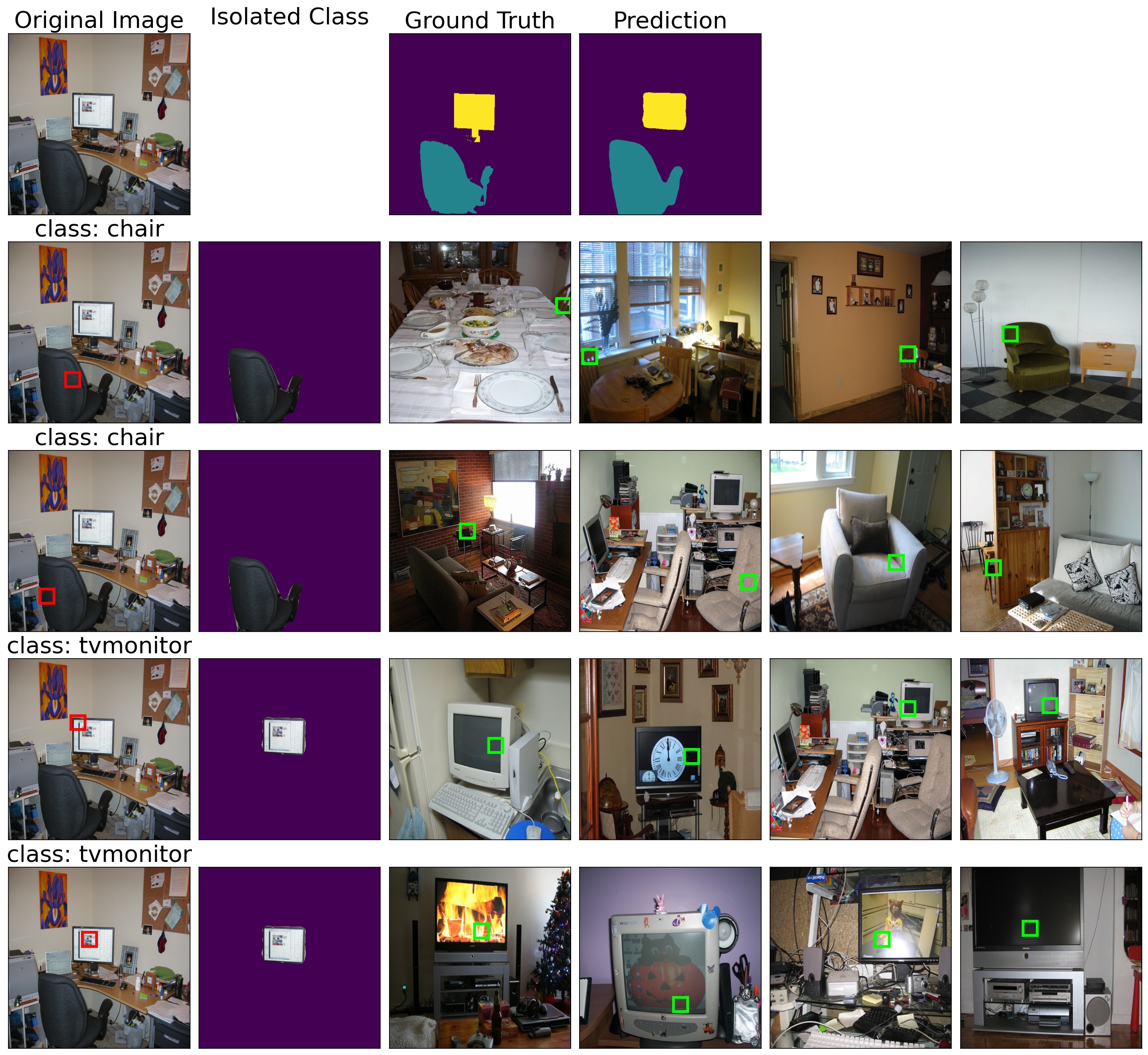}
    \caption{Visual explanation for the 'chair' and 'tvmonitor' objects identified during semantic segmentation}
    \label{fig:example-PASCAL5}
\end{figure*}

\begin{figure*}[t!] 
    \centering
    \includegraphics[width=0.8\textwidth]{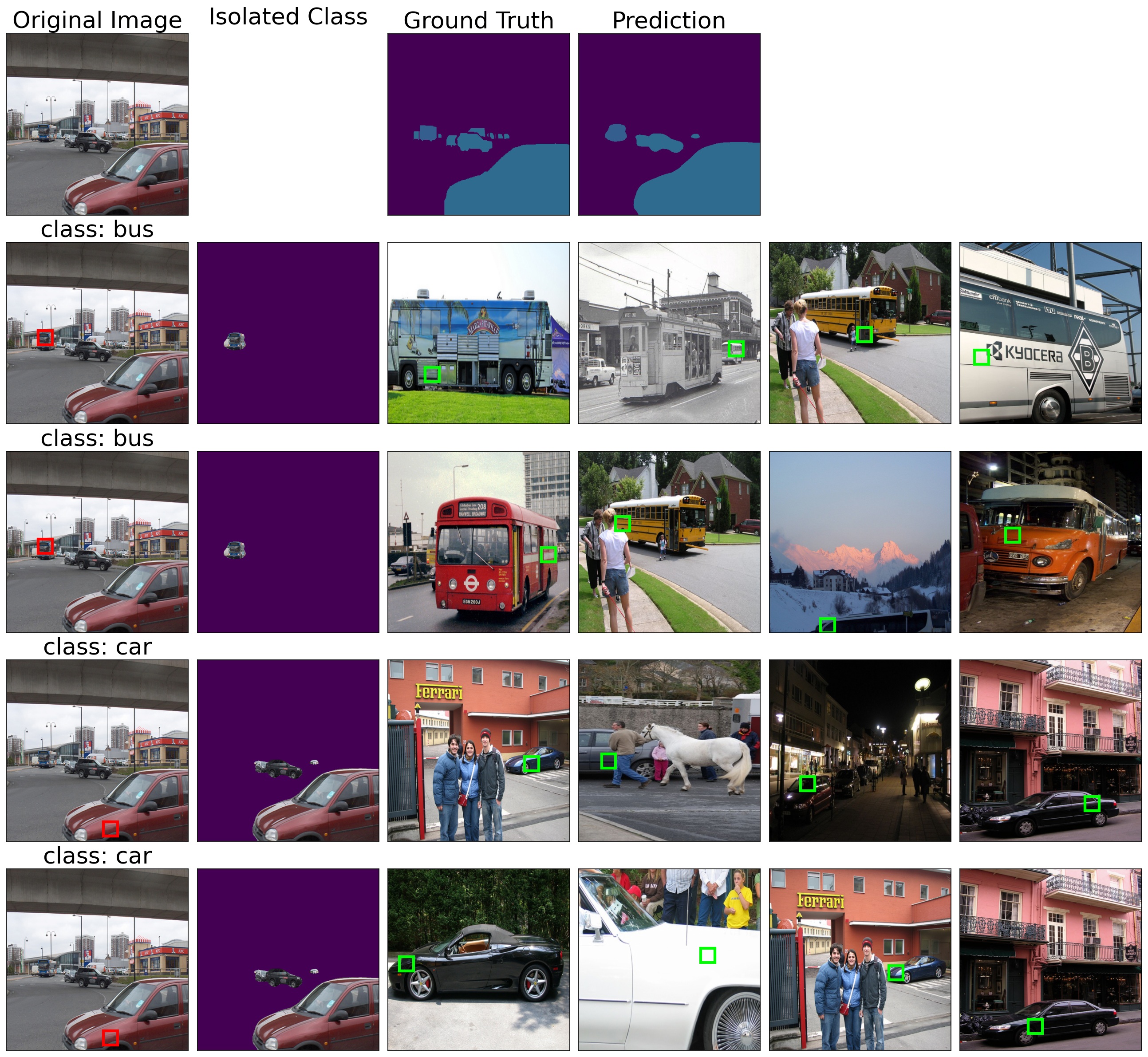}
    \caption{Visual explanation for the 'bus' and 'car' objects identified during semantic segmentation}
    \label{fig:example-PASCAL6}
\end{figure*}

\begin{figure*}[t!] 
    \centering
    \includegraphics[width=0.8\textwidth]{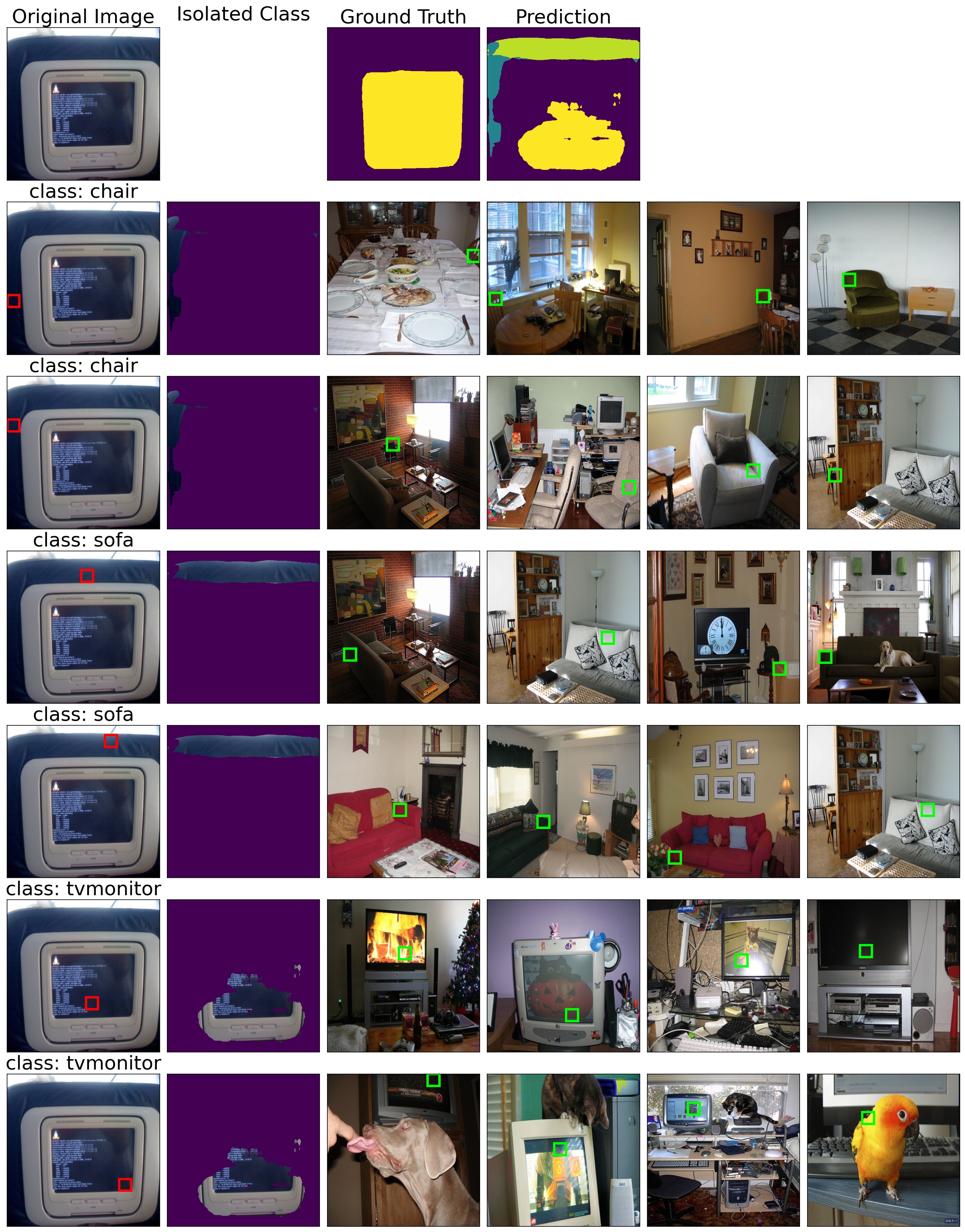}
    \caption{Visual explanation for the 'chair', 'sofa' and 'tvmonitor' objects identified during semantic segmentation}
    \label{fig:example-PASCAL7}
\end{figure*}

\begin{figure*}[t!] 
    \centering
    \includegraphics[width=0.8\textwidth]{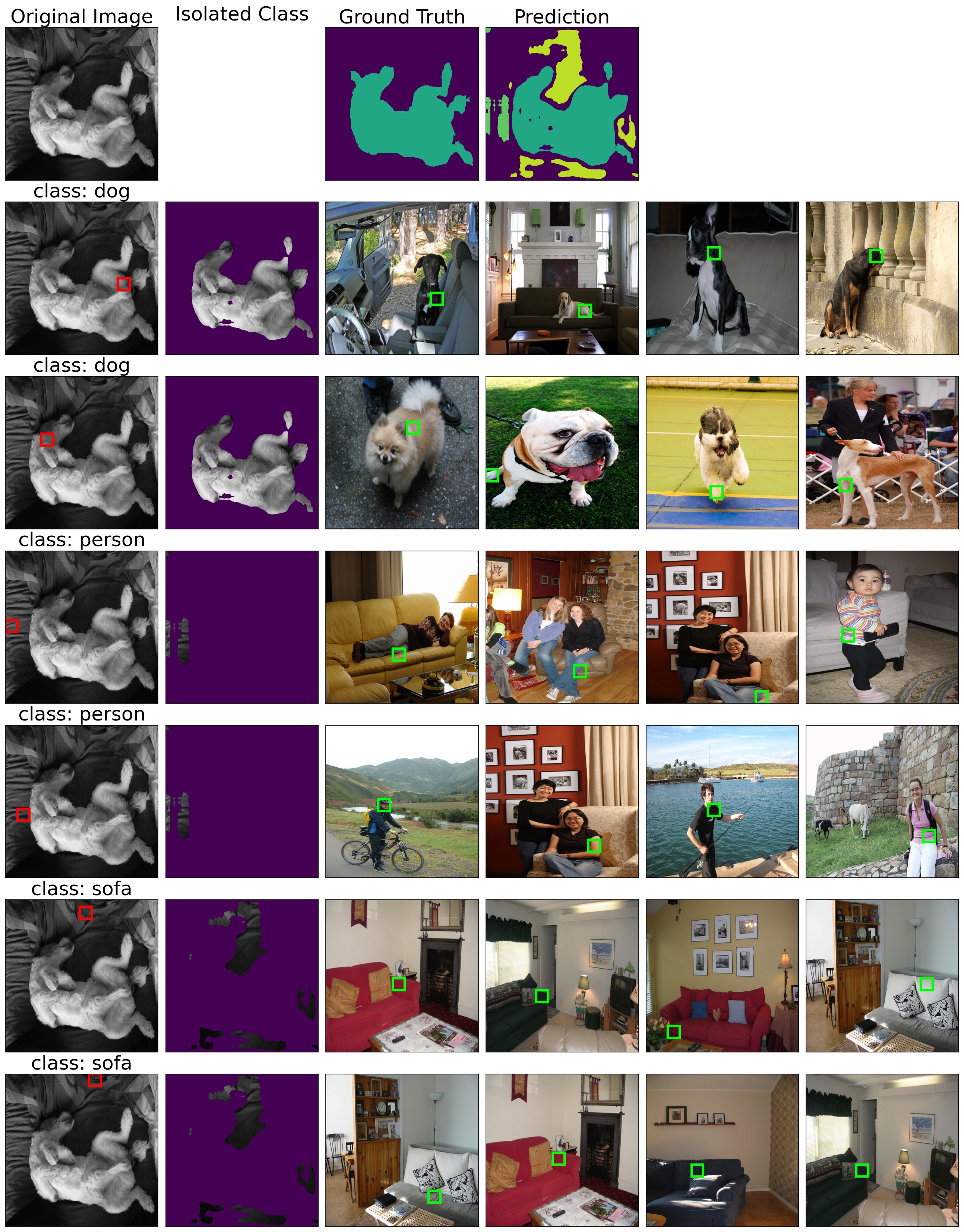}
    \caption{Visual explanation for the 'dog', 'person' and 'sofa' objects identified during semantic segmentation}
    \label{fig:example-PASCAL8}
\end{figure*}

\begin{figure*}[t!] 
    \centering
    \includegraphics[width=0.8\textwidth]{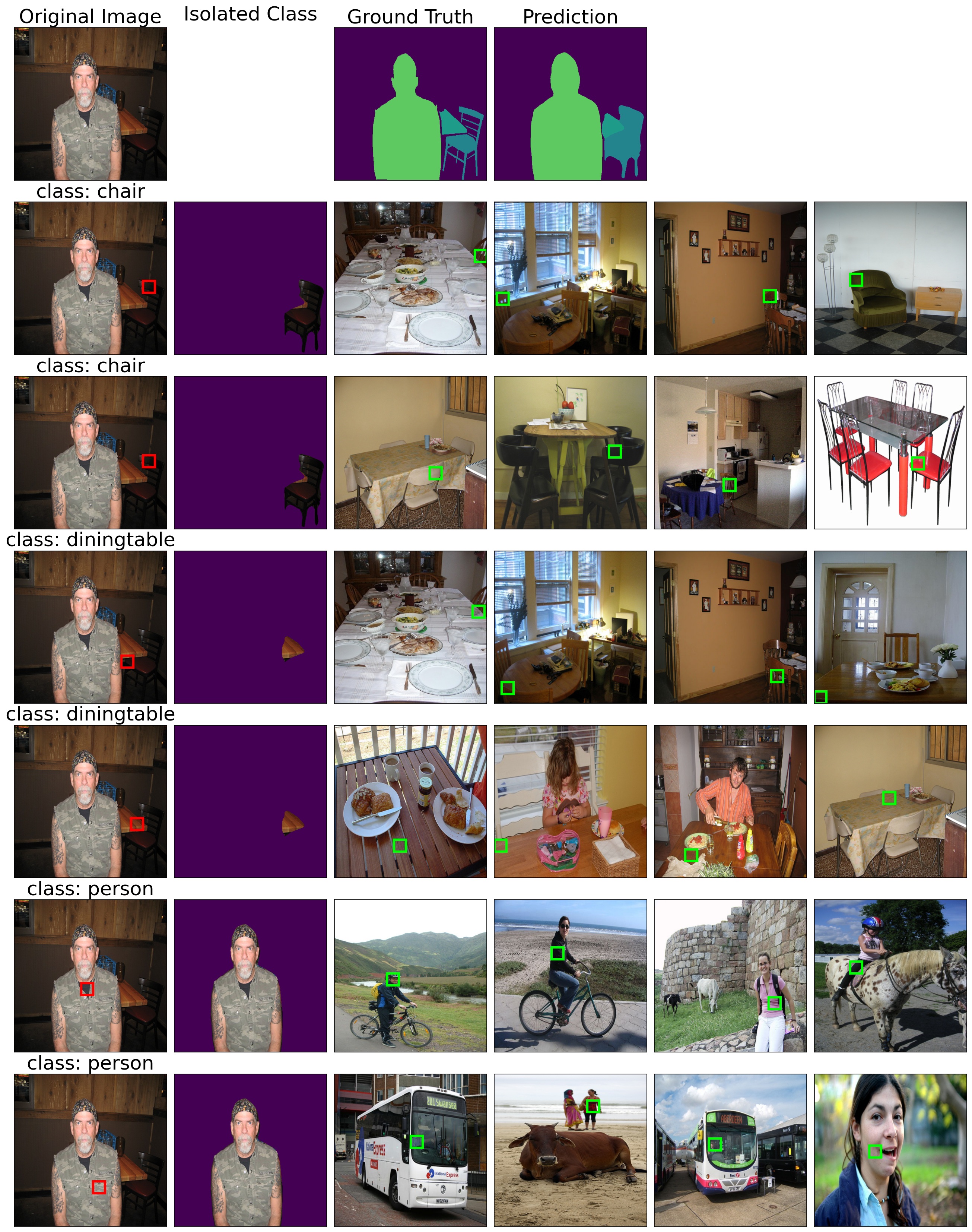}
    \caption{Visual explanation for the 'chair', 'diningtable' and 'tvmonitor' objects identified during semantic segmentation}
    \label{fig:example-PASCAL9}
\end{figure*}

Figures~\ref{fig:example-PASCAL1} through~\ref{fig:example-PASCAL9} present qualitative examples demonstrating how our method explains predictions for various object classes from the PASCAL VOC 2012 dataset. All visualizations share a consistent layout, where the top row displays the original image, the ground truth, and the model-generated segmentation. Furthermore, the first column presents the original image, the second shows the target segment to be explained, and the subsequent columns illustrate the generated prototypes (explanations).

\subsection{Point Cloud Experiments}

The point cloud adaptation of \our{} was implemented on top of the Point Transformer V1 segmentation architecture provided by the Point-BERT framework. The backbone consists of 12 transformer layers with a latent embedding dimension of 384 channels. To ensure a fair comparison with the original model, all experiments were initialized using the publicly available pre-trained Point-BERT weights, while the backbone parameters remained completely frozen throughout training.

The disentanglement module was optimized using the same procedure as described and is attached to the last layer (output layer) of point transformer. The optimization was performed for 40 epochs. Following the dynamic exemplar mining strategy, each latent channel was initially associated with the top 40 representative prototypes, with this number linearly reduced to 5 prototypes per channel by the end of training. The prototype assignments were recomputed every two epochs based on the current exemplar scores.

The semantic segmentation head follows the original Point-BERT implementation. Specifically, the patch-level representations produced by the final Point Transformer layer are propagated back to the original point cloud using PointNet++ feature propagation layers before point-wise classification. Since \our{} is inserted only as an invertible transformation of the latent representation and the subsequent classifier weights are rotated accordingly, no architectural modifications or retraining of the segmentation model are required.

Following the experimental protocol of Point-BERT, each point cloud was uniformly subsampled to 2,048 points prior to processing, and the disentanglement module was trained using a batch size of 32.

\subsection{Point Cloud examples}

\begin{figure*}[t!] 
    \centering
    \includegraphics[width=0.6\textwidth]{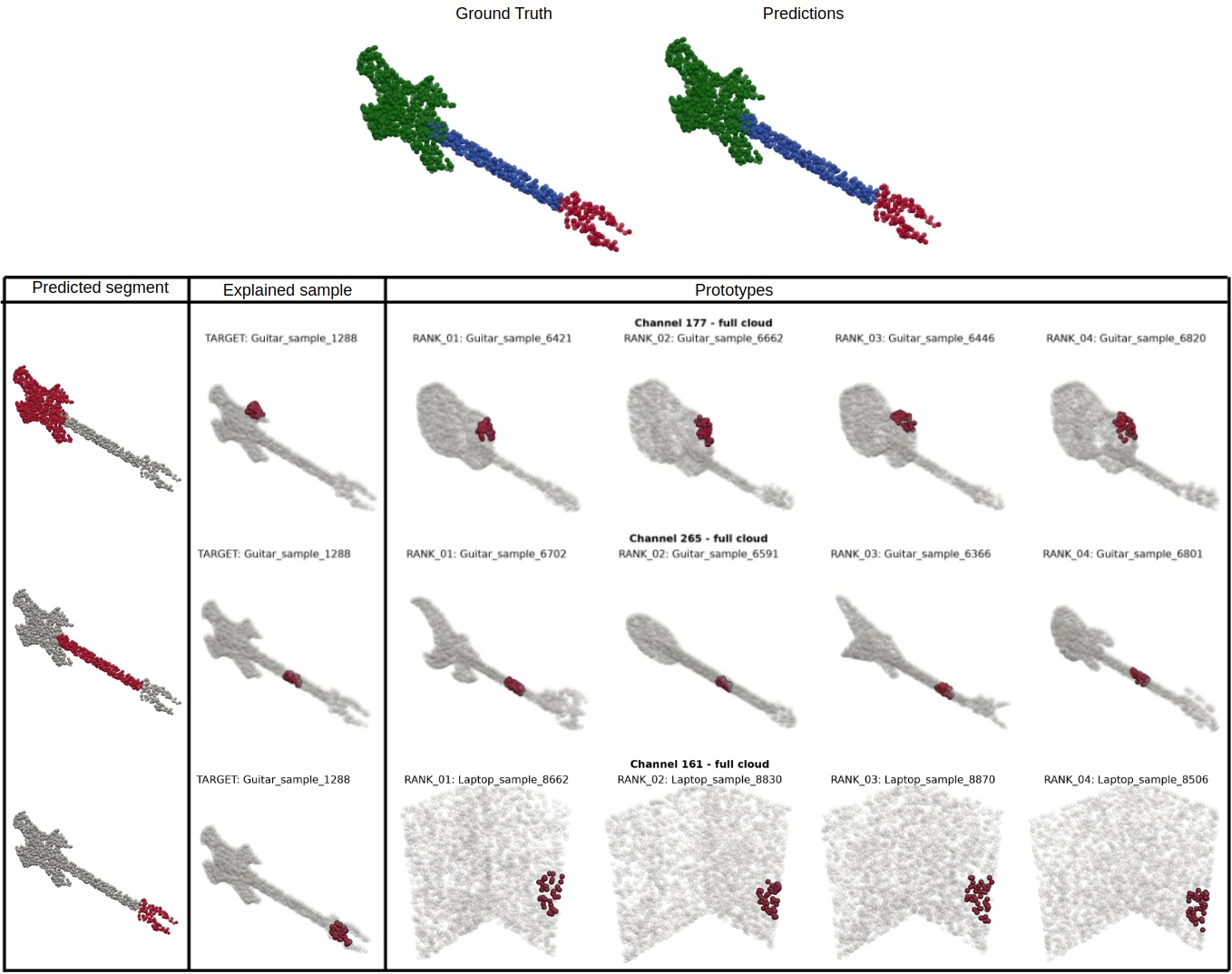}
    \caption{Explanation of the guitar object from the ShapeNetPart dataset}
    \label{fig:example-GUITAR}
\end{figure*}

\begin{figure*}[t!] 
    \centering
    \includegraphics[width=0.6\textwidth]{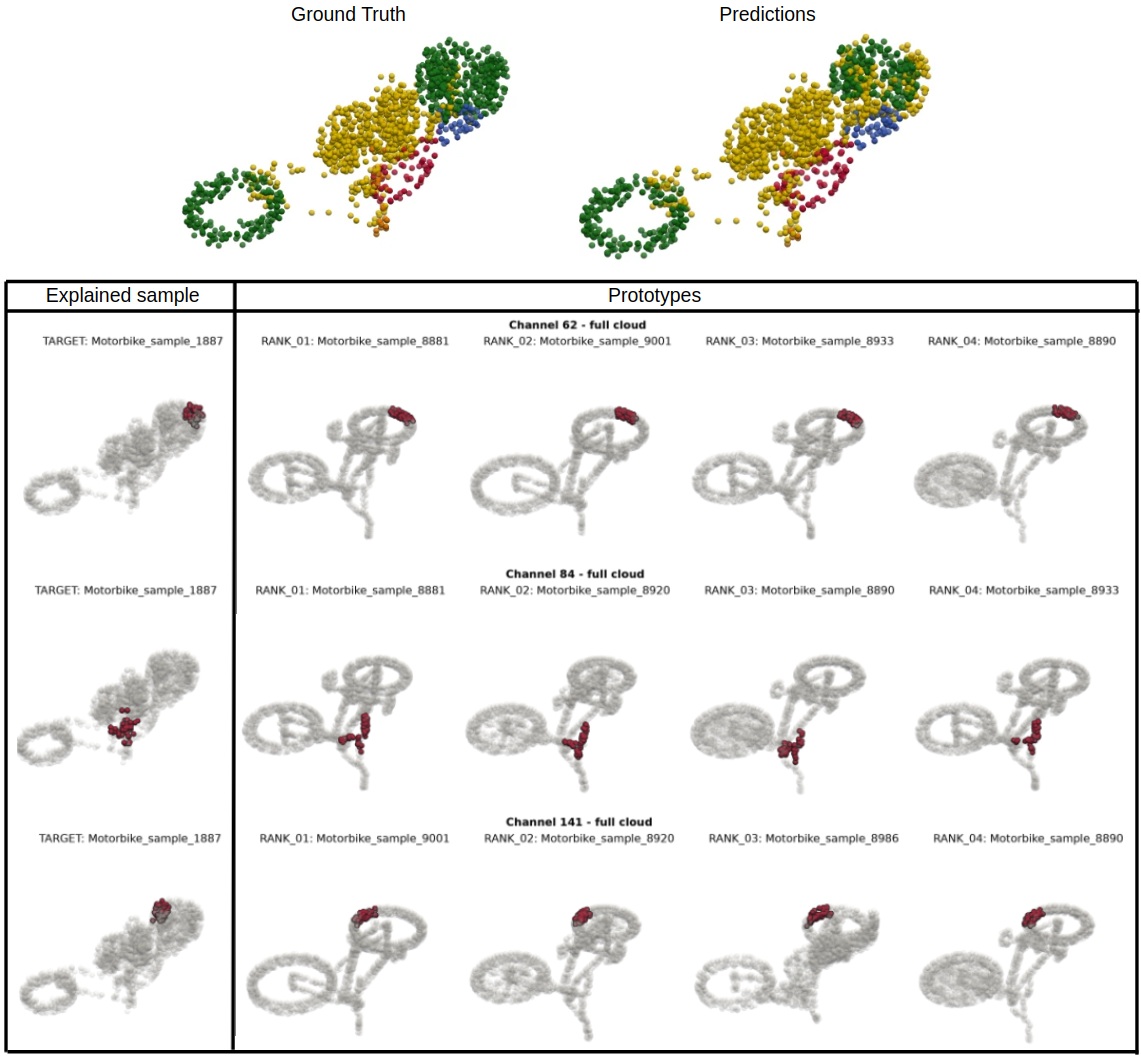}
    \caption{Explanation of the motor object from the ShapeNetPart dataset}
    \label{fig:example-MOTOR}
\end{figure*}

Figures from ~\ref{fig:example-GUITAR} to \ref{fig:example-CAP} present qualitative examples demonstrating how our method explains predictions for different object classes from the ShapeNetPart dataset. All visualizations follow the same layout. The top row shows the ground-truth part segmentation together with the predicted segmentation. The first column presents the explained object, highlighting the selected patch corresponding to the most active channel associated with a particular part class. The remaining columns display the most similar prototypes retrieved from different objects, which also exhibit high activation for the corresponding channels.

\begin{figure*}[t!] 
    \centering
    \includegraphics[width=0.6\textwidth]{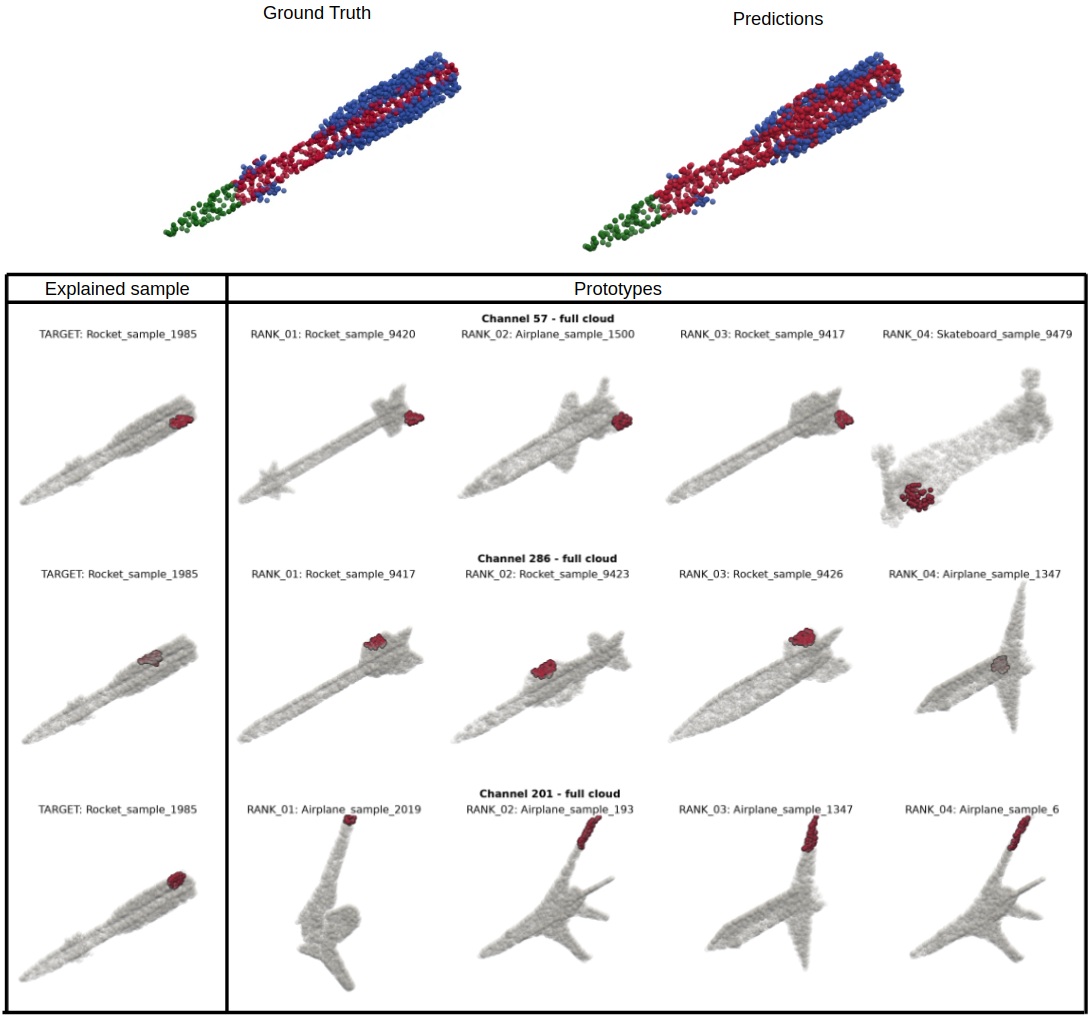}
    \caption{Explanation of the rocket object from the ShapeNetPart dataset}
    \label{fig:example-ROCKET}
\end{figure*}

\begin{figure*}[t!] 
    \centering
    \includegraphics[width=0.6\textwidth]{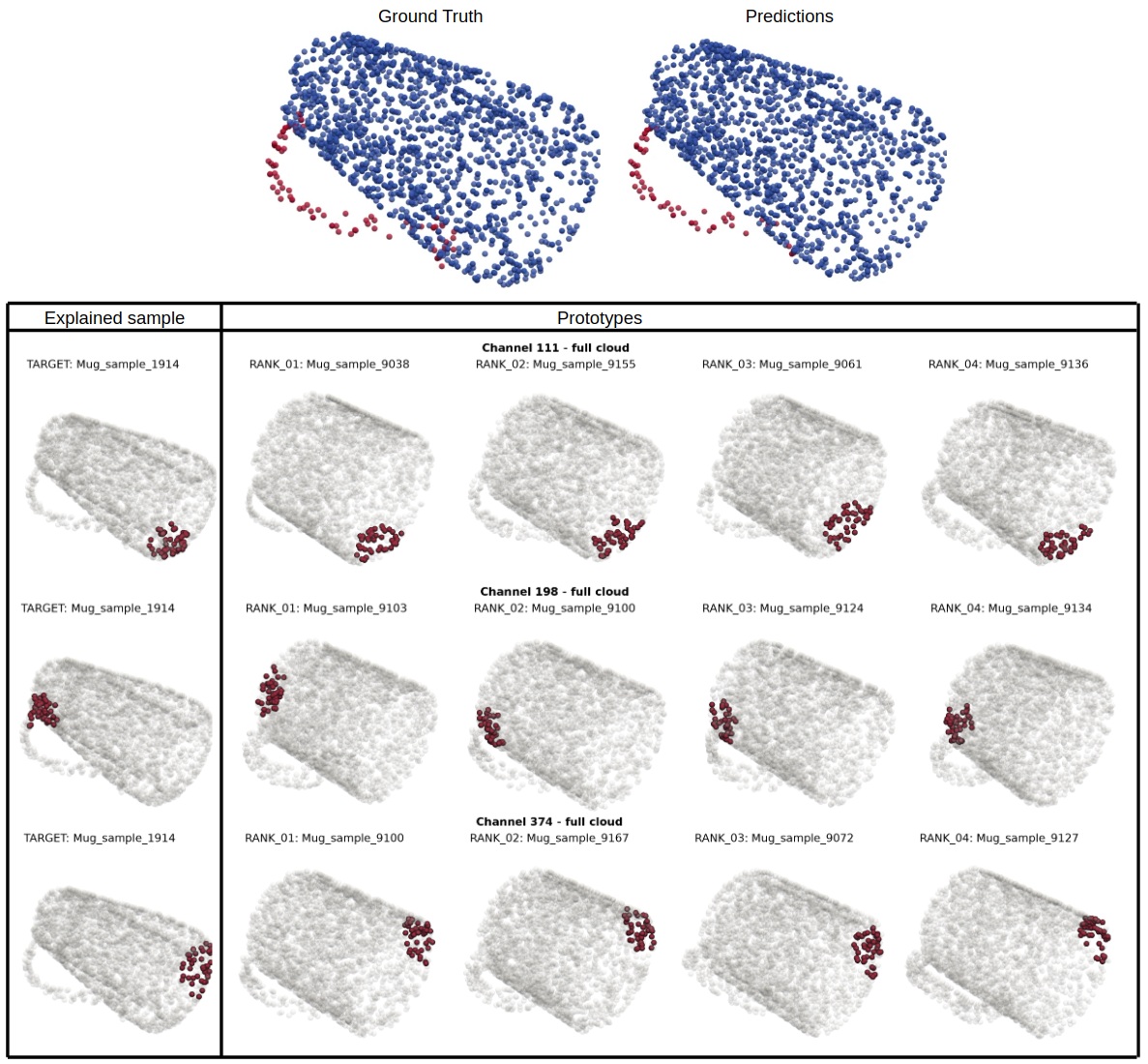}
    \caption{Explanation of the mug object from the ShapeNetPart dataset}
    \label{fig:example-ROCKET}
\end{figure*}

\begin{figure*}[t!] 
    \centering
    \includegraphics[width=0.6\textwidth]{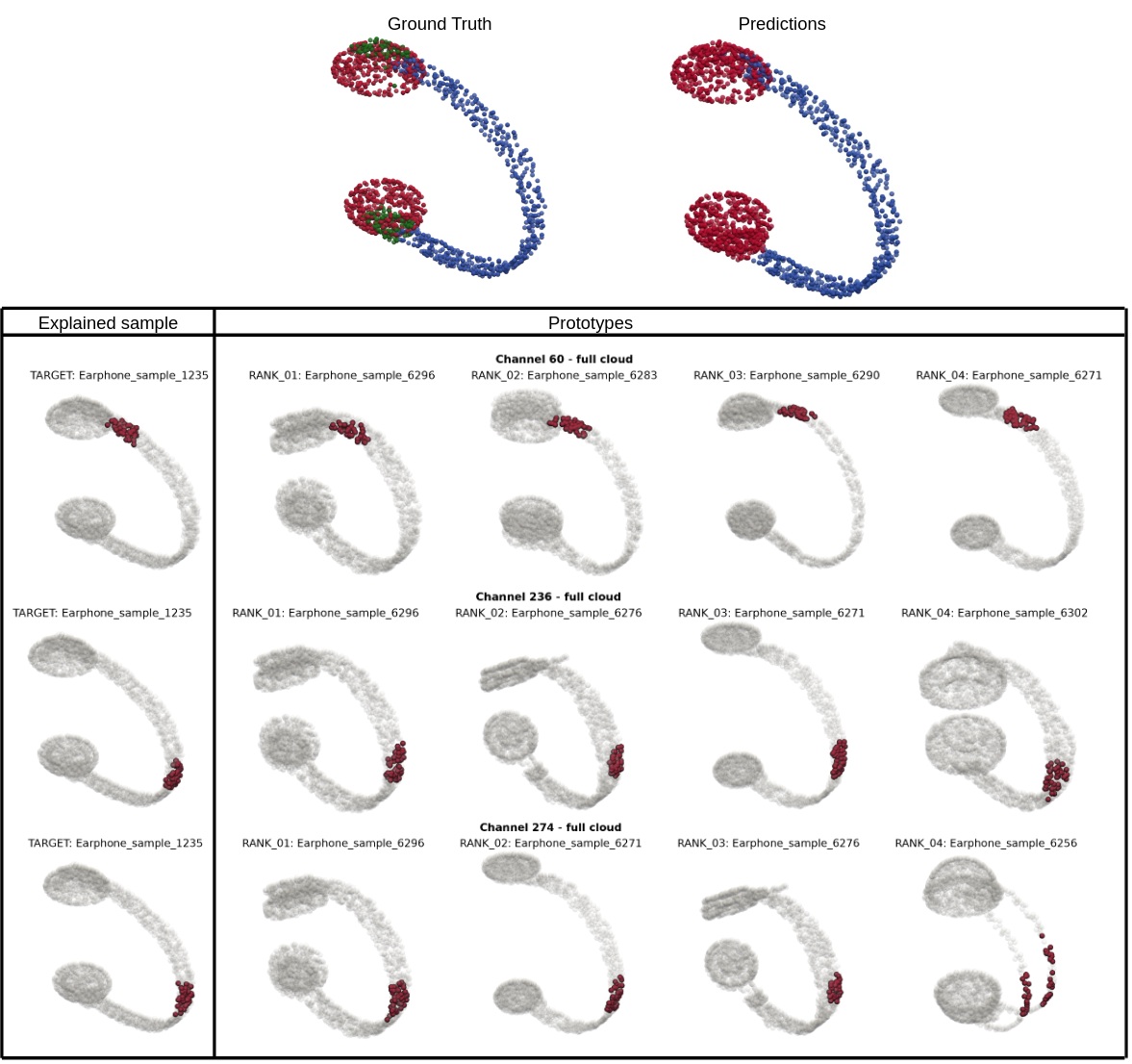}
    \caption{Explanation of the earphone object from the ShapeNetPart dataset}
    \label{fig:example-ROCKET}
\end{figure*}

\begin{figure*}[t!] 
    \centering
    \includegraphics[width=0.6\textwidth]{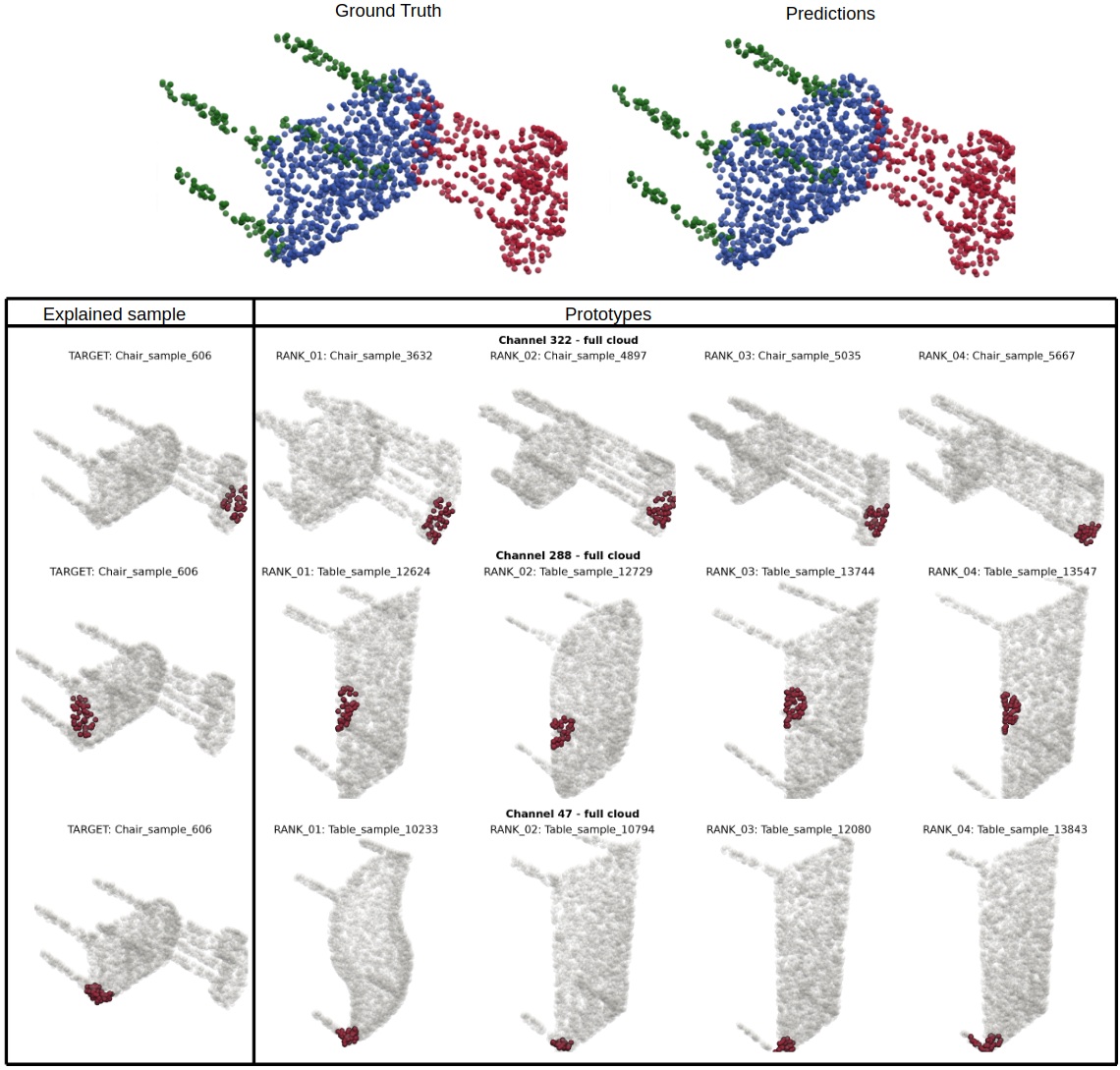}
    \caption{Explanation of the chair object from the ShapeNetPart dataset}
    \label{fig:example-ROCKET}
\end{figure*}

\begin{figure*}[t!] 
    \centering
    \includegraphics[width=0.6\textwidth]{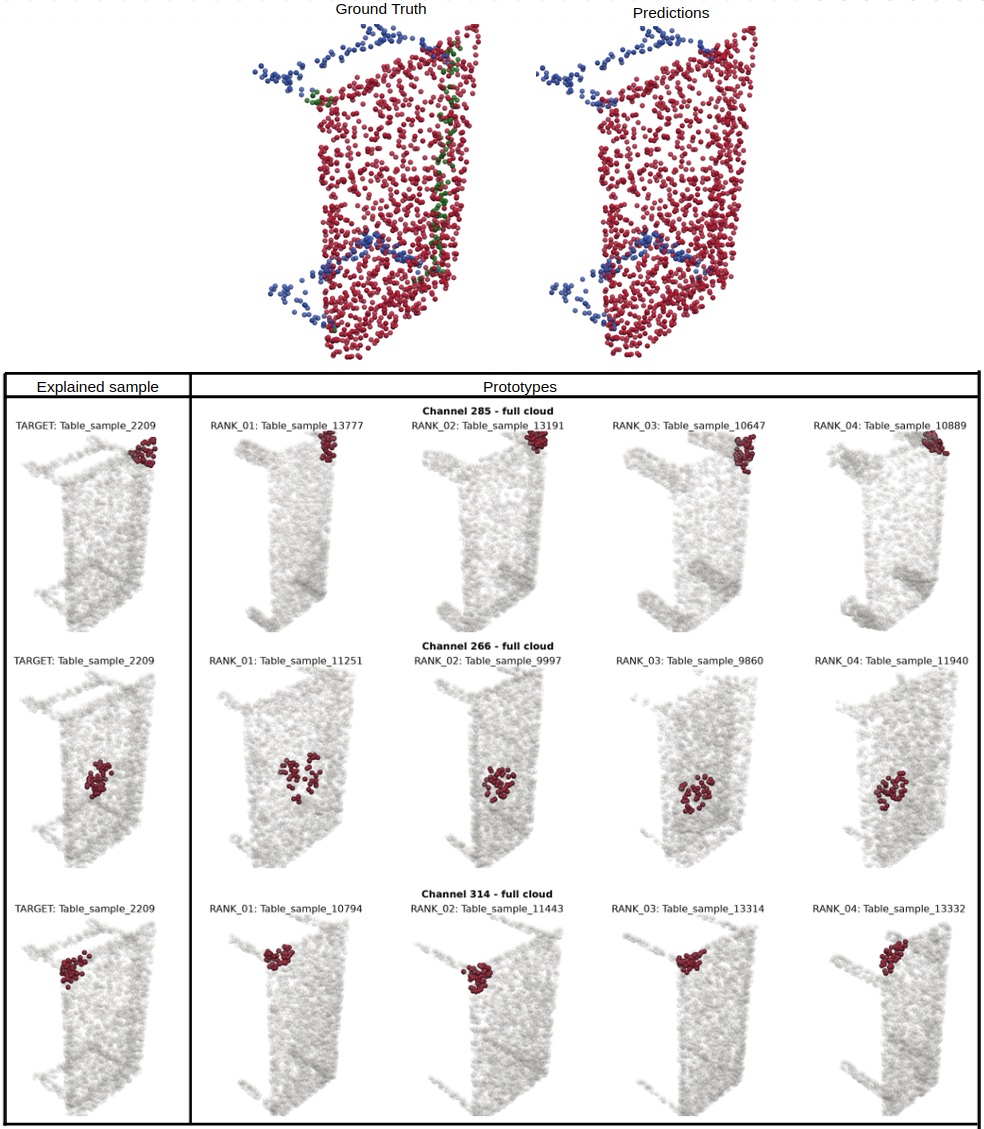}
    \caption{Explanation of the table object from the ShapeNetPart dataset}
    \label{fig:example-ROCKET}
\end{figure*}

\begin{figure*}[t!] 
    \centering
    \includegraphics[width=0.6\textwidth]{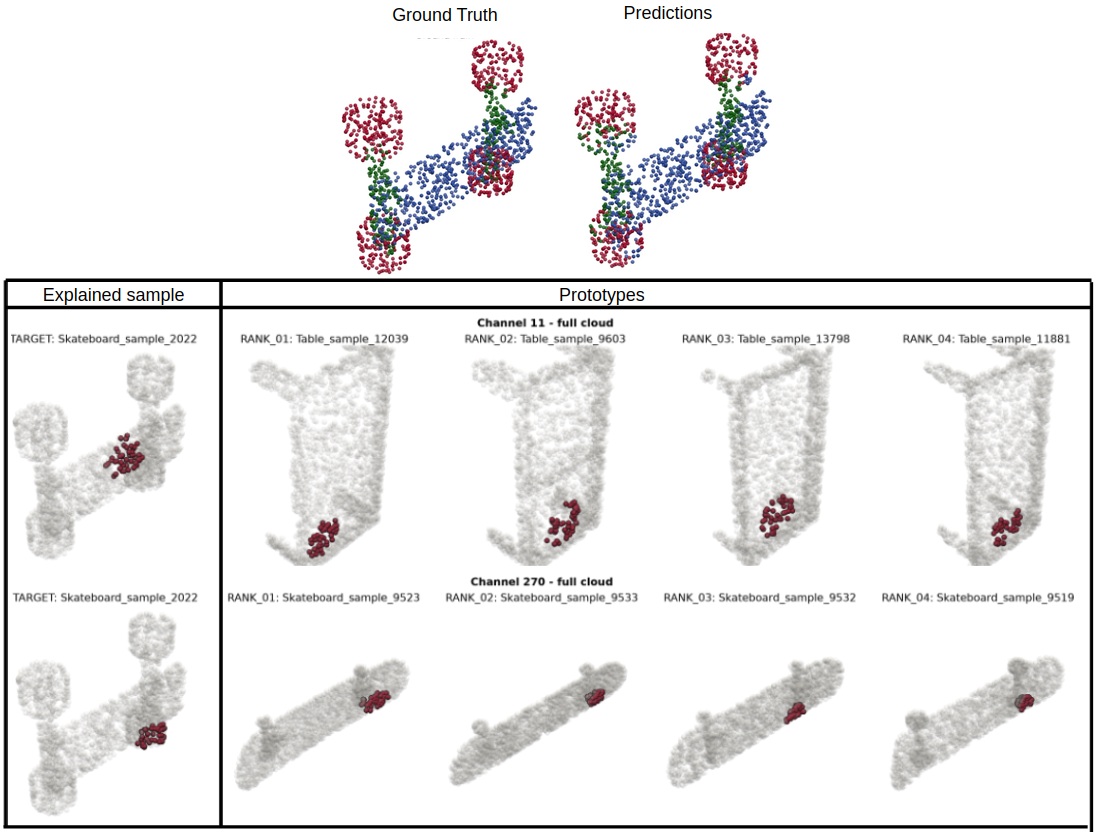}
    \caption{Explanation of the skateboard object from the ShapeNetPart dataset}
    \label{fig:example-SKATEBOARD}
\end{figure*}

\begin{figure*}[t!] 
    \centering
    \includegraphics[width=0.6\textwidth]{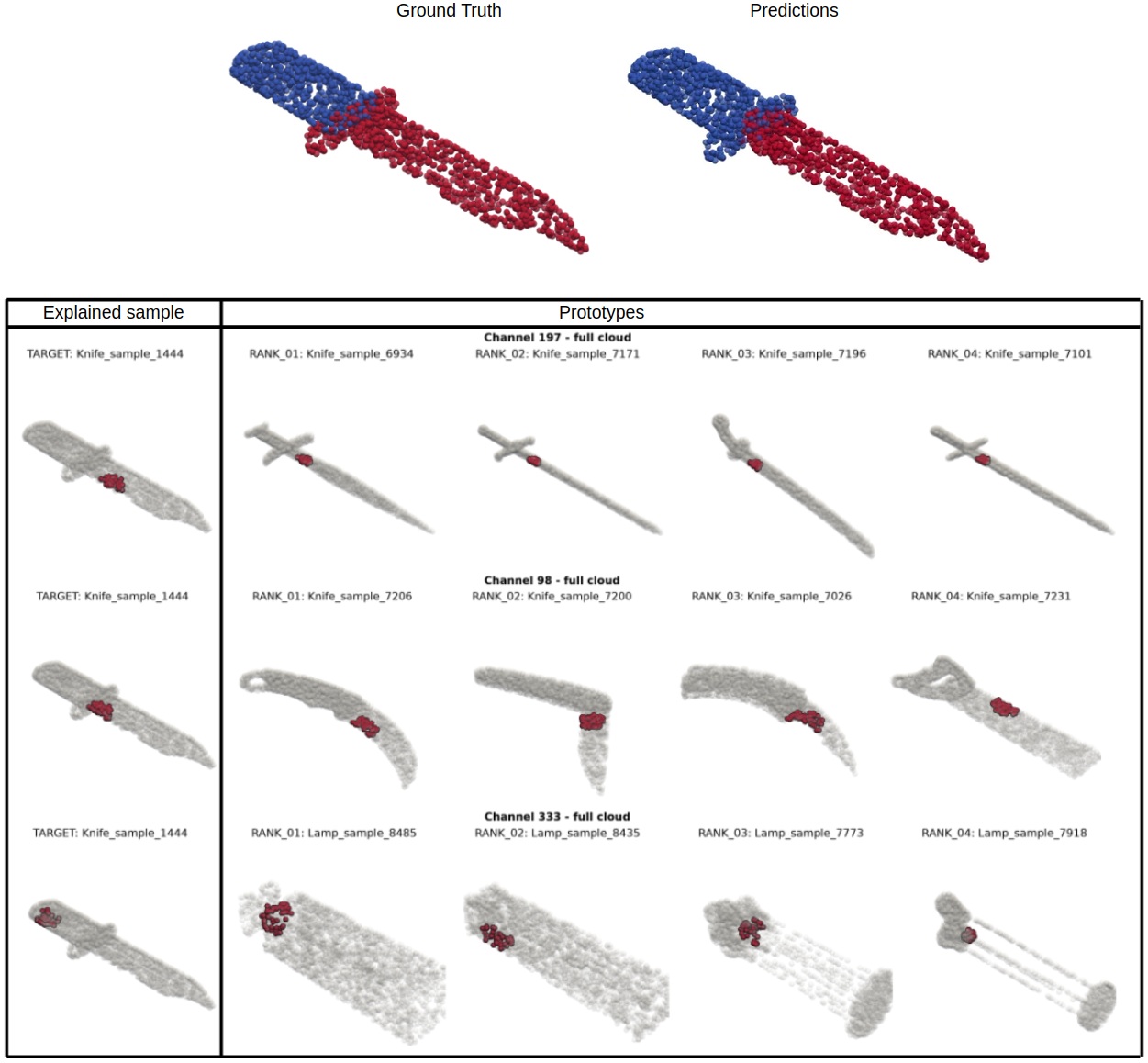}
    \caption{Explanation of the knife object from the ShapeNetPart dataset}
    \label{fig:example-KNIFE}
\end{figure*}

\begin{figure*}[t!] 
    \centering
    \includegraphics[width=0.6\textwidth]{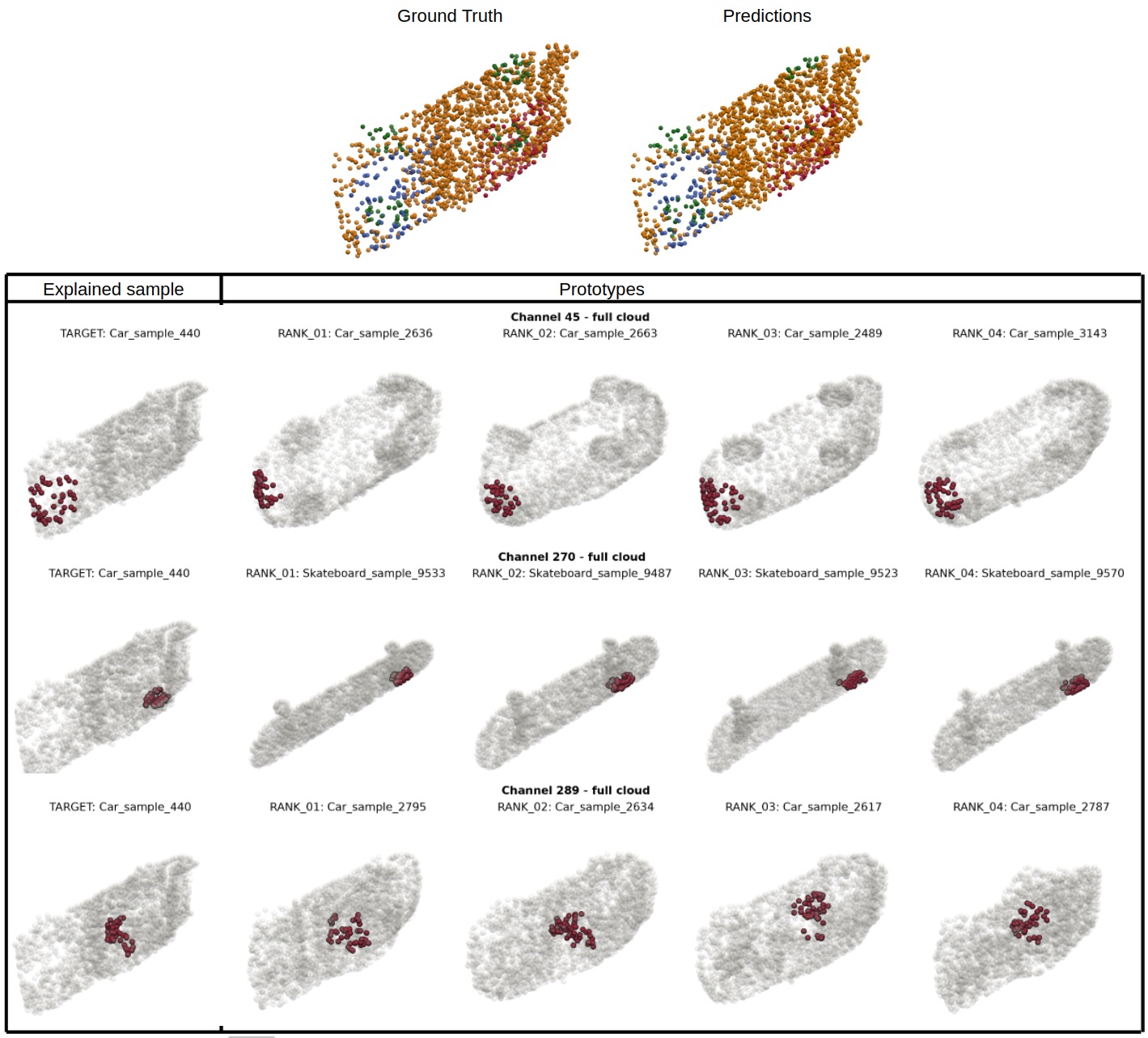}
    \caption{Explanation of the car object from the ShapeNetPart dataset}
    \label{fig:example-KNIFE}
\end{figure*}

\begin{figure*}[t!] 
    \centering
    \includegraphics[width=0.6\textwidth]{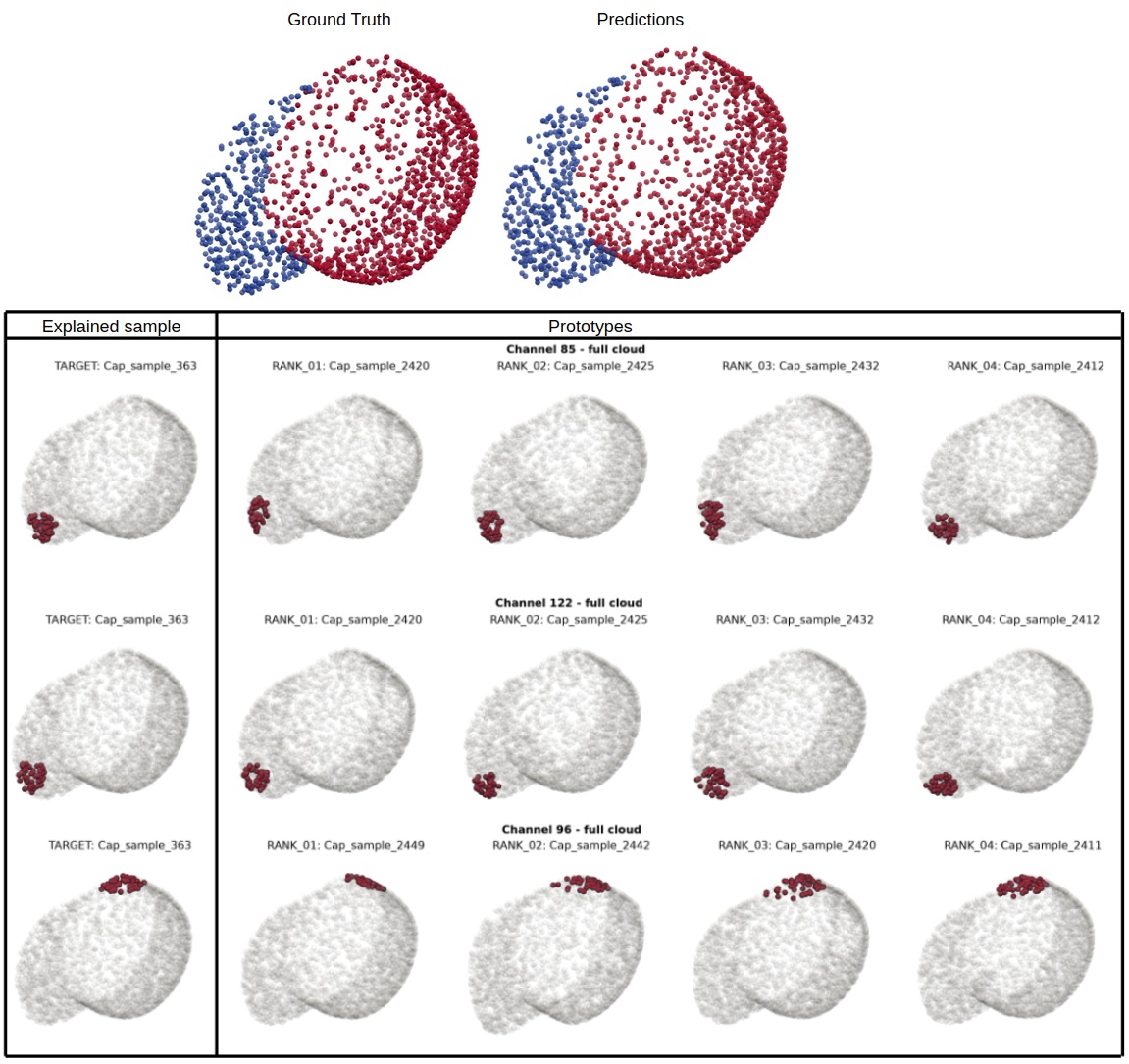}
    \caption{Explanation of the cap object from the ShapeNetPart dataset}
    \label{fig:example-CAP}
\end{figure*}
\subsection{Explaining model prediction}
After completing the optimization of the parameter matrix $A$ and aligning the orthogonal coordinate space $M$, the next step is to explain the model's pixel-level predictions for a given input image. Unlike image-level classification methods that rely on global pooling, dense semantic segmentation demands that explanations be spatially bounded to the exact predicted class boundaries to avoid background contamination. 

For an input image $I$, we restrict the feature evaluation strictly to the spatial mask of the predicted class $c_{pred}$. We then identify the peak spatial activation coordinates within the segment to extract the most relevant localized visual patches.


\begin{figure*}[t!] 
    \centering
    \includegraphics[width=\textwidth, height=1\textheight, keepaspectratio]{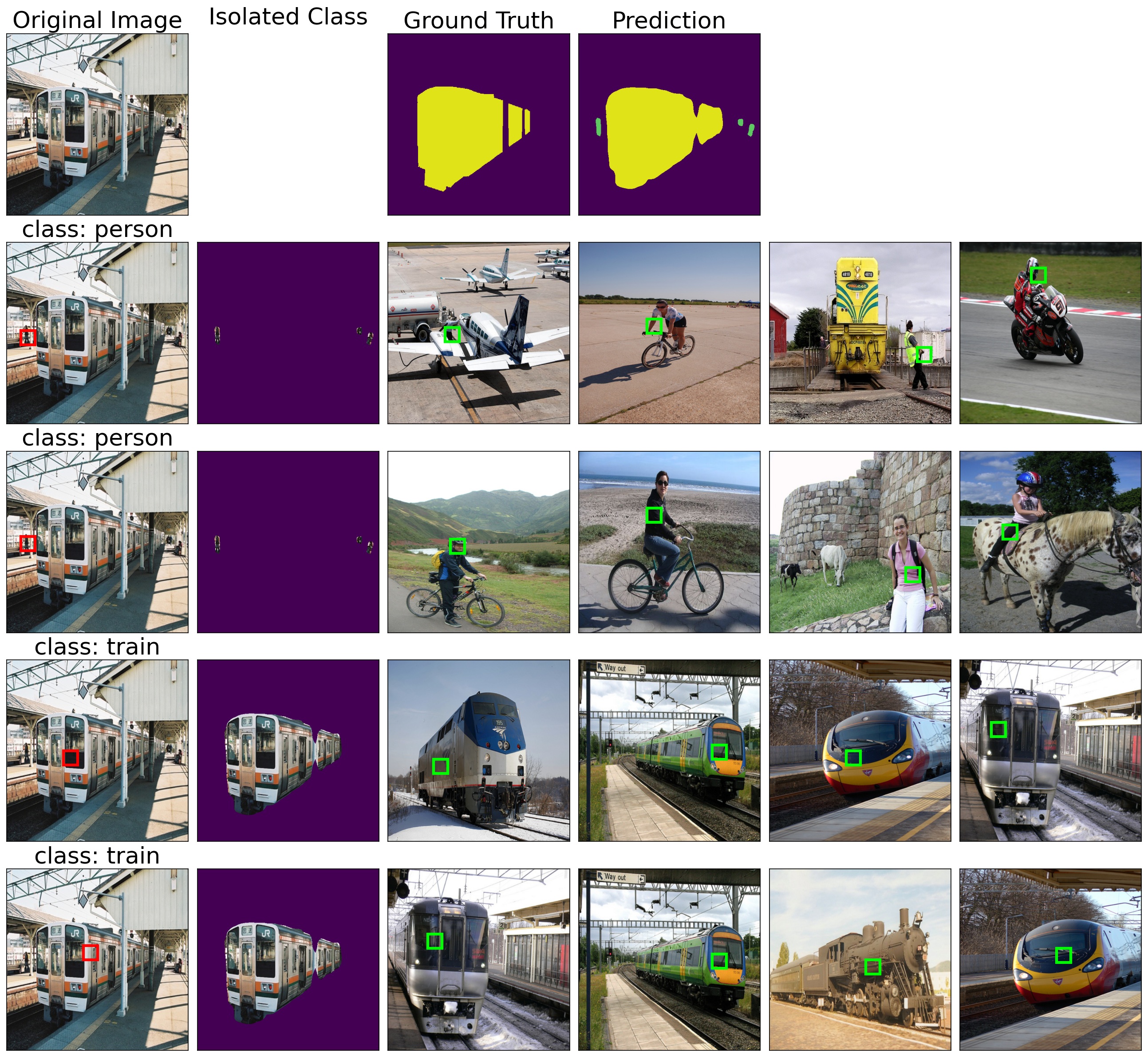}
    \caption{\textbf{An exemplary interface from the user study.} While the specific images varied across trials, all three question types shared a consistent layout featuring six visual examples. Participants evaluated the generated explanations using a 5-point Likert scale based on three distinct criteria: \textbf{(1) Visual Similarity:} ``To what extent is the presented prototype similar to the elements visible in the input image?'' (1 -- completely dissimilar, 2 -- rather dissimilar, 3 -- partially similar, 4 -- similar, 5 -- very similar). \textbf{(2) Visual Coherence:} ``To what extent do the prototypes in the same row represent a visually coherent idea or theme?'' (1 -- completely inconsistent, 2 -- rather inconsistent, 3 -- partially consistent, 4 -- consistent, 5 -- very consistent). \textbf{(3) Feature Presence:} ``Does the prototype represent a feature that can actually be observed in the input image?'' (1 -- the feature is not present, 2 -- rather not visible, 3 -- hard to tell, 4 -- rather visible, 5 -- clearly visible).}
    \label{fig:5.1userstudy}
\end{figure*}

\begin{figure*}[t!] 
    \centering
    \includegraphics[width=\textwidth, height=1\textheight, keepaspectratio]{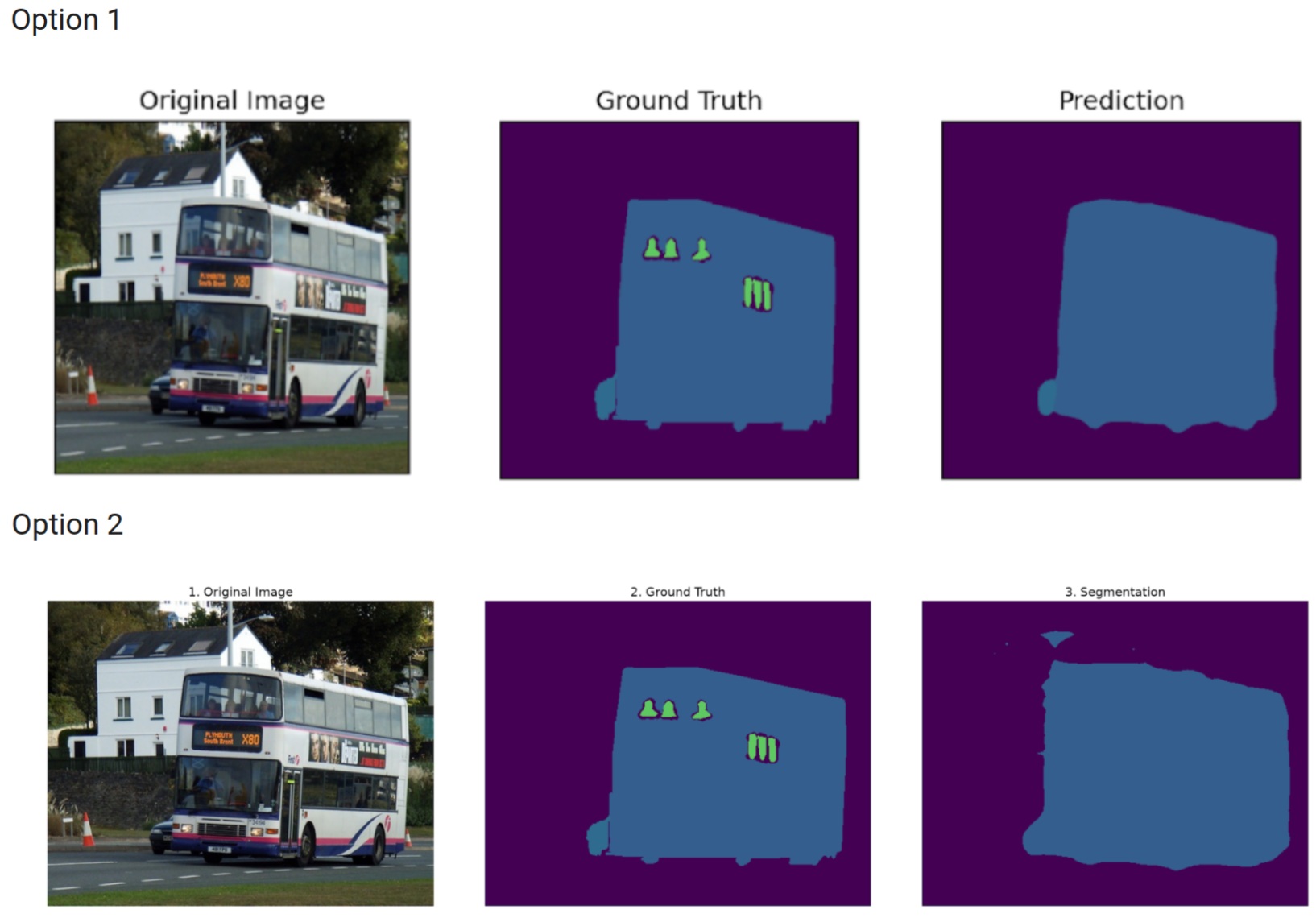}
    \caption{\textbf{Segmentation quality comparison interface.} In this task of the user study, participants were asked a question: ``Which method generates a better segmentation?''. Users compared the spatial segmentation masks produced by \our{} against those generated by the ante-hoc ScaledProtoSeg method. For this comparison, participants could explicitly choose their preferred segmentation or select a neutral ``Neither is better'' option. Notably, SegGradCam was deliberately excluded from this specific task; because both \our{} and SegGradCam are strictly post-hoc methods operating on the identical frozen DeepLabV3 backbone, they inherently yield the exact same categorical segmentation outputs. Thus, this task effectively evaluated the visual quality of our perfectly preserved baseline predictions against the altered masks produced by the ante-hoc architecture.}
    \label{fig:5.2userstudy}
\end{figure*}

\end{document}